\documentclass[10pt,journal,compsoc]{IEEEtran}
\ifCLASSOPTIONcompsoc
  \usepackage[nocompress]{cite}
\else
  \usepackage{cite}
\fi
\ifCLASSINFOpdf
\else
\fi
\usepackage{graphicx}
\usepackage{amsmath,amssymb} 
\usepackage{color}
\usepackage{multirow}
\usepackage{xspace}
\usepackage{algorithm}
\usepackage{algorithmicx}
\usepackage{algpseudocode}
\usepackage{lineno}
\usepackage{bbding}
\usepackage{ragged2e}
\usepackage{amsfonts}
\usepackage{url}
\usepackage{booktabs}
\usepackage{threeparttable}
\usepackage{tabularx}
\usepackage{caption}
\usepackage{array}

\renewcommand{\raggedright}{\leftskip=0pt \rightskip=0pt plus 0cm}

\floatname{algorithm}{Algorithm}

\makeatletter
\DeclareRobustCommand\onedot{\futurelet\@let@token\@onedot}
\def\@onedot{\ifx\@let@token.\else.\null\fi\xspace}

\def\eg{{e.g}\onedot} 
\def\ie{{i.e}\onedot} 
 
\def\etc{{etc}\onedot} \def\vs{{vs}\onedot}
\def\wrt{w.r.t\onedot} 
\def\etal{{et al}\onedot}

\makeatother

\begin{document}
%
\title{Mixed Supervised Object Detection \\ with Robust Objectness Transfer}
%
%
%
%

\author{Yan~Li,~\IEEEmembership{Student~Member,~IEEE,}
        Junge~Zhang,~\IEEEmembership{Member,~IEEE,}\\
        Kaiqi~Huang*,~\IEEEmembership{Senior~Member,~IEEE,}
        Jianguo~Zhang,~\IEEEmembership{Senior~Member,~IEEE}
\IEEEcompsocitemizethanks{
\IEEEcompsocthanksitem Y.~Li, and J.~Zhang are with the Center for Research on Intelligent Perception and Computing (CRIPAC), National Laboratory of Pattern Recognition (NLPR) of Institute of Automation, Chinese Academy of Sciences (CASIA), Beijing 100190, China, and also with the University of Chinese Academy of Sciences (UCAS), Beijing 100049, China.

E-mail: yan.li@cripac.ia.ac.cn, jgzhang@nlpr.ia.ac.cn.

\IEEEcompsocthanksitem K. ~Huang is with the Center for Research on Intelligent Perception and Computing (CRIPAC), National Laboratory of Pattern Recognition (NLPR) of Institute of Automation, Chinese Academy of Sciences (CASIA), Beijing 100190, China, the University of Chinese Academy of Sciences (UCAS), Beijing 100049, China, and also with the CAS Center for Excellence in Brain Science and Intelligence Technology, Shanghai 200031, P.R.China. E-mail: kqhuang@nlpr.ia.ac.cn

\IEEEcompsocthanksitem Jianguo~Zhang is with the Computing, School of Science and Engineering, University of Dundee, Dundee DD1 4HN, United Kingdom.

E-mail: j.n.zhang@dundee.ac.uk.
}
\thanks{Manuscript received 5 July, 2017; revised 12 Jan. 2018; accepted 24 Feb. 2018.}
\thanks{Data of publication 0.0000; data of current version 0.0000.}
\thanks{(Corresponding author: Kaiqi Huang.)}
\thanks{Recommended for acceptance by A. Vedaldi.}
\thanks{For information on obtaining reprints of this article, please send e-mail to: reprints@ieee.org, and reference the Digital Object Identifier below.}
\thanks{Digital Object Identifier no. 10.1109\/TPAMI.2018.2810288}
}
%
%

\markboth{Journal of \LaTeX\ Class Files,~Vol.~XX, No.~XX, XX~2018}%
{Li \MakeLowercase{\textit{et al.}}: Mixed Supervised Object Detection with Robust Objectness Transfer}
%



\IEEEtitleabstractindextext{%
\begin{abstract} In this paper, we consider the problem of leveraging existing fully labeled categories to improve the weakly supervised detection (WSD) of new object categories, which we refer to as mixed supervised detection (MSD). Different from previous MSD methods that directly transfer the pre-trained object detectors from existing categories to new categories, we propose a more reasonable and robust objectness transfer approach for MSD. In our framework, we first learn domain-invariant objectness knowledge from the existing fully labeled categories. The knowledge is modeled based on invariant features that are robust to the distribution discrepancy between the existing categories and new categories; therefore the resulting knowledge would generalize well to new categories and could assist detection models to reject distractors (\eg, object parts) in weakly labeled images of new categories. Under the guidance of learned objectness knowledge, we utilize multiple instance learning (MIL) to model the concepts of both objects and distractors and to further improve the ability of rejecting distractors in weakly labeled images. Our robust objectness transfer approach outperforms the existing MSD methods, and achieves state-of-the-art results on the challenging ILSVRC2013 detection dataset and the PASCAL VOC datasets.
\end{abstract}
}

\maketitle

\IEEEdisplaynontitleabstractindextext

%
\IEEEpeerreviewmaketitle

\IEEEraisesectionheading{\section{Introduction}\label{sec:intro}}

%
%
%
%
\IEEEPARstart{R}{ecently}, object detection has been improved drastically in performance and scale with the development of convolutional neural networks (CNNs) \cite{gidaris2015object,girshick2015fast,girshick2014rich,ren2015faster,dai2016r} and the introduction of benchmarking detection datasets (\eg, PASCAL VOC \cite{everingham2005pascal}, MS COCO \cite{lin2014microsoft} and ILSVRC2013 detection dataset \cite{russakovsky2015imagenet}). The supervised training process of state-of-the-art object detectors requires a large number of fully labeled images with bounding box annotations. However, the bounding box annotations are very difficult to acquire; thus supervised training of a high-performance detector at such a scale is not feasible, when we need to deal with hundreds of thousands of new objects in real-world applications.

\begin{figure}[t]
\begin{center}
   \includegraphics[width=0.95\linewidth]{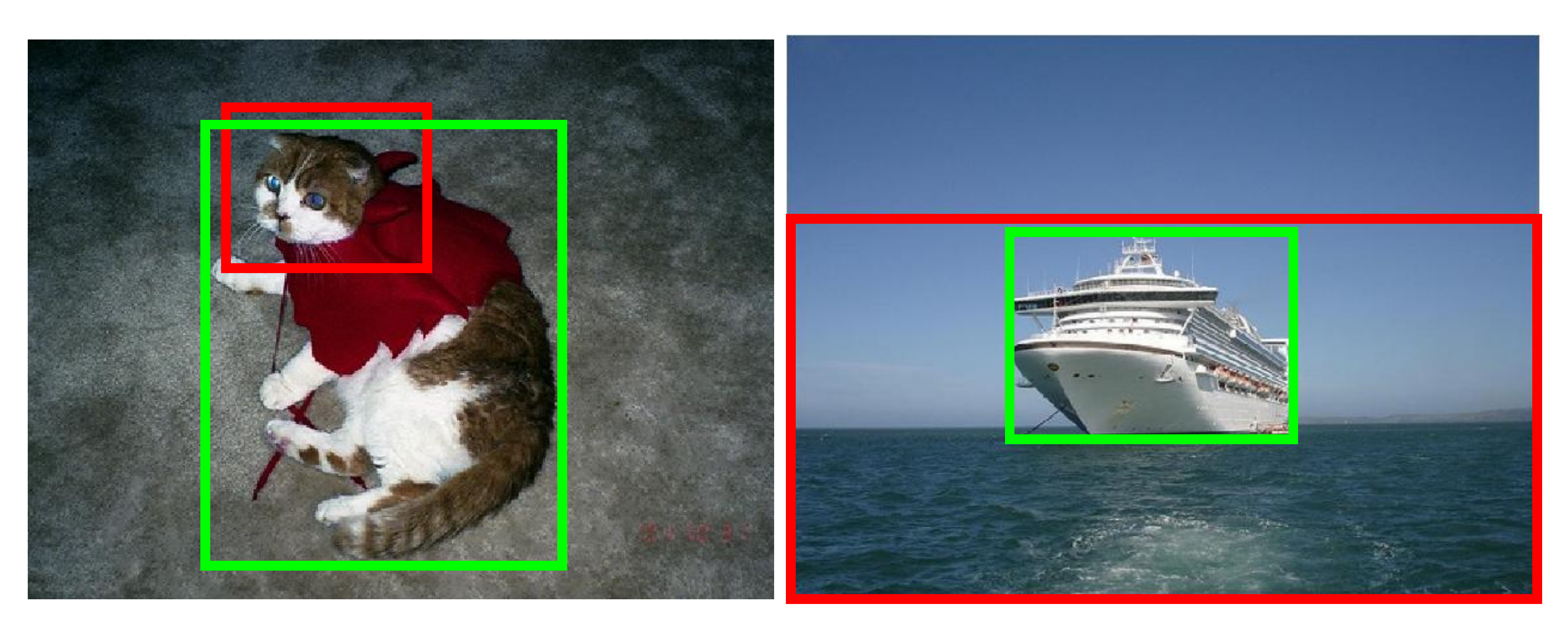}
\end{center}
   \caption[The LOF caption]{Failure examples of weakly supervised detection\footnotemark. The weakly supervised detection tends to confuse the objects (\textit{cat, boat}) with the co-occurring distractors (\textit{cat face, water}).}
\label{fig:fail}
\end{figure}

On the other hand, weakly labeled data (\ie, object categories with image labels only) are relatively easier to obtain either manually or through search engine. The problem of learning a detector on images with such weak labels is called \textit{weakly} supervised detection (WSD). It is often formulated as a multiple instance learning (MIL) problem \cite{dietterich1997solving}, in which each image is used as a bag of regions. However, unfortunately, the clear definition of \textit{instances} for object regions is missing in such formulation. As illustrated in Fig. \ref{fig:fail}, WSD tends to confuse the objects with co-occurring distractors (context or object part) \cite{bilen2014weakly,bilen2015weakly,bilen2016weakly,cinbis2014multi,cinbis2017weakly,kolesnikov2016improving,song2014learning}. Those co-occurring distractors are in principle perfectly \textit{valid} positive instances considering the image class labels only, though not the actual object being sought, for example \textit{boat} versus \textit{boat with water} and ``\textit{whole cat}'' versus ``\textit{cat face}''. The key issue in weakly supervised setting is that WSD trains an object detector by optimizing objective functions for an image-level classification instead of the region-level detection. Considering a typical case when both ``\textit{whole cat}'' region and ``\textit{cat face}'' region present in many positive \textit{cat} images but do not appear in negative images, WSD cannot distinguish the two regions and is also likely to select ``\textit{cat face}'' region as the desired \textit{cat} region, as either of them could distinguish \textit{cat} from other categories. Such an incapability of distinguishing the objects from distractors leads to lots of false detections and limits the detection performance. This pitfall is inherent in current weakly supervised setting.

However we argue that the pitfall could be better addressed by building detectors over a mixed set of images with strong labels (\ie, bounding box annotations) and weak labels (\ie, image-level labels); we call such a problem \textit{mixed} supervised detection (MSD). The mixed supervised learning offers two key advantages: 1) limiting the amount of annotations due to the use of weak labels; 2) leveraging fully labeled public datasets to assist training on weak labels. For simplicity and clarity, in the following we term the object categories with bounding box annotations as \textit{strong} categories, while the categories with image-level annotations only are called \textit{weak} categories. We would like to highlight that, different from \textit{semi}-supervised detection \cite{misra2015watch,rosenberg2005semi,yang2013semi}, strong categories in mixed supervised detection have \textit{NO} overlap with weak categories, \ie, objects in the weak categories are novel categories \wrt the strong categories. The trained detector is expected to detect an object instance from one of those novel categories in an unseen image. Thus MSD is a more challenging problem. The existing MSD methods \cite{hoffman2014lsda,hoffman2015detector,rochan2015weakly,tang2016large} utilize a straightforward pipeline that directly transfers the object detectors learned on strong categories to weak categories following some hand-crafted strategies. We argue that a better approach to solve MSD should be capable of 1) learning strong domain-invariant knowledge from strong categories and 2) robustly transferring the learned knowledge to weak categories.

\renewcommand{\thefootnote}{\fnsymbol{footnote}}
\footnotetext{The results are obtained using our implementation of WSDDN \cite{bilen2016weakly}.}

In this paper, we propose a robust objectness transfer approach for mixed supervised learning. In our approach, we first aim to learn domain-invariant objectness knowledge from strong categories with CNN models. The objectness knowledge learned from annotated boxes of strong categories could facilitate rejecting the distractors in weakly labeled images. Meanwhile, the unlabeled regions from weak categories are utilized to make the learned objectness generalize well to new categories. Concretely, we cast the learning of objectness as a domain adaptation problem, considering the strong categories as our source domain, and the weak categories as our target domain. The objectness model is trained with the embedding representations where both strong and weak categories are indistinguishable. Thus the learned objectness will be invariant to the change of domains (from strong to weak categories). The domain-invariant objectness has two important characteristics: 1) category independent, to generalize well to unseen categories and 2) object sensitive, to reliably reject distractors in weak categories. We believe that such knowledge is appropriate for mixed supervised detection scenarios. After that, the objectness knowledge is applied to separate the objects and distractors in weak categories and a simple way is to use the separated objects as pseudo ground truths to train object detectors on weak categories. However, we believe that the difference between objects and distractors in weak categories can be better modeled by a further learning process. Specifically, we consider the separated objects and distractors as ``object bag'' and ``distractor bag'' and aim to model the concepts of objects and distractors under a standard multiple instance learning (MIL) framework. Finally, with the improved object and distractor concepts, the detection model is capable of distinguishing the ground truth objects from distractors, and leads to much better performance.

In summary, our contributions are three-fold:
\begin{itemize}
    \item A robust objectness transfer approach is proposed for MSD. Different from previous MSD methods that directly transfer pre-trained object detectors from strong to weak categories with hand-crafted strategies, our method automatically \textit{learns} the domain-invariant knowledge by incorporating weak categories into the knowledge learning process.
\item  We design a MIL-based framework to further model the difference between objects and distractors and to improve the ability of reliably rejecting distractors in weakly labeled images.
    \item The proposed method outperforms both the state-of-the-art MSD methods and the baselines on the challenging ILSVRC2013 detection and PASCAL VOC 2007, 2010, 2012 datasets.
\end{itemize}

\section{Related work \label{sec:rework}}

\subsection{Weakly Supervised Detection}

To reduce the annotation cost in object detection, weakly supervised detection (WSD) methods \cite{bilen2014weakly,bilen2015weakly,bilen2016weakly,cinbis2014multi,cinbis2017weakly,diba2016weakly,kantorov2016contextlocnet,kolesnikov2016improving,li2016weakly,song2014learning,wang2014weakly} attempt to learn object detectors using only image category labels. In weakly supervised setting, the optimization of WSD methods is an image-level classification instead of the required region-level detection, thus the WSD methods tend to select distractors (local optima) and their performance strongly depends on the initialization. Song \etal \cite{song2014learning} and Wang \etal \cite{wang2014weakly} use clustering method to obtain better initializations. Cinbis \etal \cite{cinbis2014multi,cinbis2017weakly} propose a multi-fold training strategy of MIL to avoid the local optima: the dataset is split into 10 subsets. When selecting high-score proposals from a subset, the detectors trained on other subsets are used. Bilen \etal \cite{bilen2014weakly} propose a smoothed version of MIL where soft labels are related to the region proposals instead of choosing the ones with highest confidence. WSDDN \cite{bilen2016weakly} utilizes a two-stream architecture to train the recognition model and to select the discriminative regions in parallel to avoid using the recognition model itself to select high confident regions, which is able to relieve the local optima phenomenon. Based on WSDDN, Kantorov \etal \cite{kantorov2016contextlocnet} propose to utilize the context information to reject distractors and obtain more reliable detections. While these approaches are promising, the local optima problem has not yet been solved. The performance of WSD methods is still far from acceptable.

\subsection{Mixed Supervised Detection}
To learn well-performing object detectors with image category labels, several methods aim to utilize fully labeled data of different categories (strong categories) to improve the detection performance on the weakly labeled categories (weak categories), which is referred to as mixed supervised detection (MSD). Shi \etal \cite{shi2017transfer} propose to learn a rank model on strong categories based on the appearance similarity. Then the rank model is transferred to weak categories to select the top-ranked regions as objects. Guillaumin and Ferrari \cite{guillaumin2012large} conduct the MSD on ImageNet \cite{russakovsky2015imagenet}. By exploiting the semantic hierarchy of ImageNet, the key idea in \cite{guillaumin2012large} is to localize objects of a weak category by transferring knowledge from its \textit{ancestor} and \textit{sibling} strong categories. Hoffman \etal \cite{hoffman2014lsda} propose a Large Scale Object Detection through Adaptation (LSDA) algorithm to address the MSD problem. In their method, the classifier and detector differences are learned on strong categories and then transferred from several ``similar" strong categories to a weak category. The weak category applies the transferred differences to adjust its classifier to corresponding object detector. In \cite{hoffman2015detector}, Hoffman \etal utilize the same strategy to adapt the intermediate representations from strong to weak categories and then solve a standard MIL problem on weak categories based on transferred representations. Recently, Tang \etal \cite{tang2016large} propose a Large Scale Semi-supervised Object Detection (LSSOD) method to improve the LSDA. LSSOD follows the same approach in LSDA but selects the ``similar" categories by considering more informed visual and semantic similarities. In LSDA-based methods, both classifiers and object detectors are trained with 8-layer AlexNet model \cite{krizhevsky2012imagenet}, and the parameters of layers 1-7 are the same for all categories in the models. Thus LSDA has to assume as a prior that the differences learned on layers 1-7 are category-invariant (note they are not \textit{learned} to achieve category-invariance as what we do in this paper), those differences learned on strong categories are directly applied to weak categories. However, when the distribution discrepancy between strong and weak categories becomes significant, this assumption is no longer valid and the detection performance will be weakened greatly. Rocha \etal \cite{rochan2015weakly} propose a method called Weakly Supervised Localization Using Appearance Transfer (WSLAT) to solve the MSD based on semantic knowledge. In WSLAT, the strong and weak categories are represented as fixed length vectors (called ``word embedding") \cite{mikolov2013distributed}. Then the object detectors can be transferred from one category to another based on their semantic relationship. However, the semantic information is still an indirect measurement of complex object categories. The transferred detectors cannot obtain good performance in their method. In contrast, our model learns more robust and transferable objectness to support the learning of WSD on weak categories, which is able to effectively relieve the impacts caused by distribution discrepancy between strong and weak categories. Recently, Shi \etal \cite{shi2017weakly} propose a new mixed supervised learning setting, where the auxiliary fully labeled annotations correspond to the pixel-level segmentation annotations and the knowledge is learned from the segmentation models.

\subsection{Objectness Knowledge}
Learning objectness knowledge to improve the detection performance has been explored by many previous works \cite{alexe2012measuring,kuo2015deepbox,uijlings2013selective,zitnick2014edge}. Most of them measure objectness with low-level cues such as saliency \cite{alexe2012measuring}, contour information \cite{zitnick2014edge} and hierarchical superpixels \cite{uijlings2013selective}. These approaches are ``class-agnostic'' and can be applied to new categories. However, the ability of these objectness knowledge is limited due to their incapabilities of capturing high-level cues. In fact, some high-level cues are also beneficial to detecting objects. For example, many animal object categories might share the same high-level structures (limbs around the body). Detecting such structures could infer to the presence of objects, but it is difficult to learn such high-level structures with low-level information only. Based on this consideration, learning objectness from annotated images with deep CNN models is likely to perform better, and such CNN-based objectness knowledge has already been explored by DeepBox \cite{kuo2015deepbox}. With a large number of annotated images, DeepBox aims to let CNN model itself figure out what low-, mid- and high-level object cues are most discriminative and it achieves promising results in fully supervised detection scenario. However, such CNN-based approach is not appropriate for our MSD task. Since the objectness in DeepBox is learned on strong categories and directly applied to different categories without considering the distribution discrepancy problem. It would decrease the performance of DeepBox on new categories, especially when the available strong categories is limited. Different from previous objectness methods, our domain-invariant deep objectness model is the first work that incorporates the domain invariance into the CNN-based objectness method, which makes it possible to obtain sufficient objectness from existing annotated categories and to transfer the knowledge to new categories well.

\begin{figure*}[htb]
\begin{center}
   \includegraphics[width=0.95\linewidth]{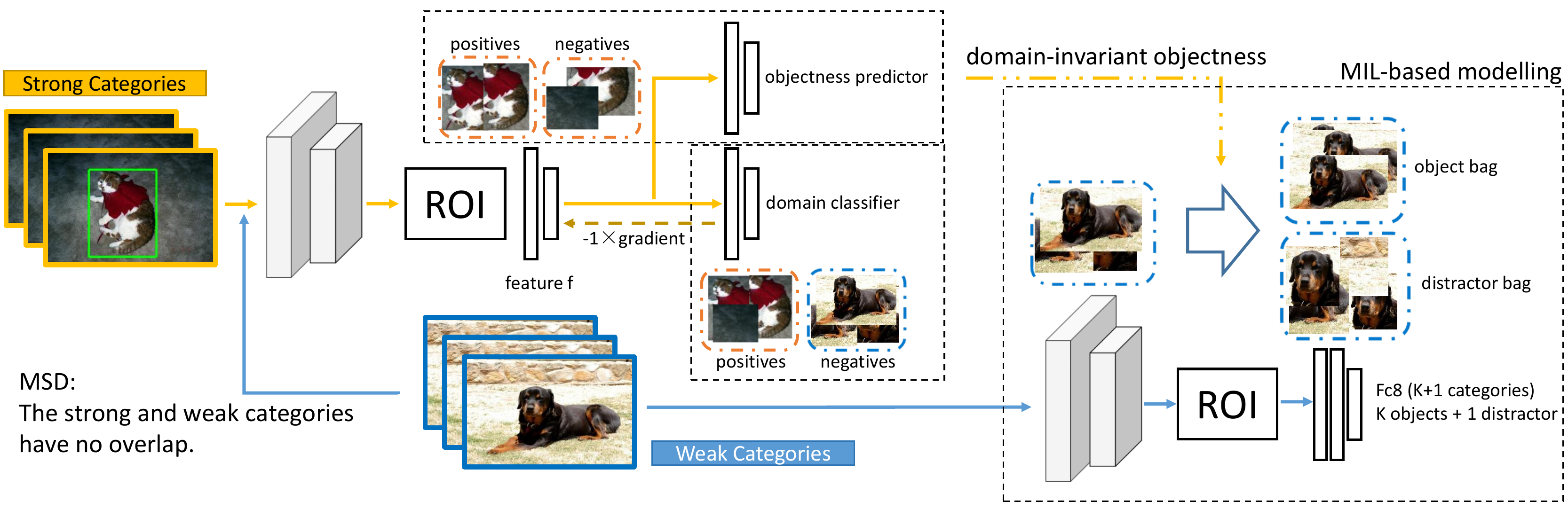}
\end{center}
\vspace{-1em}
   \caption{The proposed robust objectness transfer approach for MSD. During the learning of objectness, the annotated boxes from fully labeled categories (``strong" categories, \eg, \textit{cat}) are used to train the objectness predictor; meanwhile, the unlabeled regions from weakly labeled categories (``weak" categories, \eg, \textit{dog}) are also applied to learn a domain classifier. During training, the gradients from the domain classifier are reversed to make the feature f invariant to the change of categories. The learned objectness is firstly utilized to roughly distinguish the objects and distractors and then a MIL-based approach is used to further model the difference between the objects and distractors.}
\label{fig:framework}
\end{figure*}

\section{Task Definition}

In the mixed supervised learning case, we assume that we already have a set of fully labeled categories, which is called ``strong categories" and denoted as $\mathcal{S}$. Meanwhile we have some weakly labeled categories called ``weak categories", denoted as $\mathcal{W}$. Both bounding box annotations and image-level labels are available for set $\mathcal{S}$; for set $\mathcal{W}$, we only have access to their image category labels. In our detection scenario, the strong categories and weak categories have \textit{no} overlap. This is quite different from the ``mixed supervised learning'' explored in Cinis's work \cite{cinbis2017weakly}, which actually belongs to \textit{semi}-supervised learning where typically a small amount of fully labeled data with a large amount of weakly labeled data are provided for the same category.

\section{Method \label{sec:method}}
Our robust objectness transfer framework is illustrated in Fig. \ref{fig:framework}. We first learn domain-invariant objectness knowledge to assist the weakly supervised learning on weak categories. During the learning of objectness, the annotated boxes from strong categories are used to train the objectness predictor. Meanwhile, the unlabeled boxes from weak categories are also applied to learn a domain classifier, and the gradients from the domain classifier are reversed to achieve the domain invariance. The learned objectness is first utilized to roughly distinguish the objects and distractors and then a MIL-based approach is used to further model the difference between the objects and distractors. Finally, the detection model is able to recognize the \textit{distractor} category in addition to object categories and learn that these confused distractors are false detections.

\subsection{Learning Domain-invariant Objectness}
\label{section:objectness}
In our approach, we aim to model the objectness knowledge using CNN-based method, and the objectness model is directly trained on the bottom-up proposals that are generated by selective search \cite{uijlings2013selective}. By leveraging the bounding box annotations in strong categories, we cast the learning of objectness as a binary classification task: considering the regions that largely overlap with the ground truth boxes as the ``objects'' and the regions with smaller overlaps as ``non-objects''. We aim to let CNN models to automatically figure out beneficial cues for learning ``objectness''. During training, the images as well as a set of region proposals are fed to several convolutional layers, the RoI pooling layer \cite{girshick2015fast} and fully-connected (fc) layers. Each region $\mathbf{r}_{i}$ is finally mapped to a 256-dimensional vector $\mathbf{f}_{i} \in {\mathbb{R}^{256}}$ (\ie, \textit{feature} $\mathbf{f}$ in Fig. \ref{fig:framework}), which can be considered as the internal representations for input regions. Then the $\mathbf{f}_{i}$ is connected to two branches.

The first branch is the objectness predictor, denoted as $G_{obj}$. $G_{obj}$ consists of fc layers and predicts from $\mathbf{f}_{i}$ whether a region $\mathbf{r}_i$ is an object or not. As bounding box annotations are required to separate object regions and non-object regions in an image, only regions from set $\mathcal{S}$ are used to train the objectness predictor. The binary logistic loss can be used:

\begin{equation}
\label{equ:obj}
L_{obj}{\left( \mathbf{w} \right)} = - \frac{1}{n} \sum_{i=1}^{n} [y_i^{obj} \log(p_i) + (1-y_i^{obj}) \log(1-p_i)]
\end{equation}
where $p_i=\frac{1}{1+\exp(-G_{obj}(\mathbf{f_i}))}$ is the posterior probability that a region $\mathbf{r}_i$ belongs to ``objects''. The regions whose intersection-over-union (IoU) with any ground truth object is no less than 0.5 are considered as positive examples, \ie, $y_i^{obj}=1$; the regions that have a maximum IoU with ground truth in the interval $\left[ 0.1,0.5 \right)$ are negative examples, \ie, $y_i^{obj}=0$. Additionally, we balance the ratio of positive and negative samples in each image to 1:3 as the number of negative examples is far more than the positive ones.

Such objectness is trained on set $\mathcal{S}$ and cannot be extended to set $\mathcal{W}$ well, since the statistical distributions of categories in the two sets are different. Thus, we need to make the objectness learned on set $\mathcal{S}$ generalize well on set $\mathcal{W}$. Inspired by \cite{ganin2015unsupervised,ganin2016domain}, in our approach, we cast the learning of objectness as a domain adaptation problem where set $\mathcal{S}$ and $\mathcal{W}$ can be considered as source and target domain respectively, and the domain invariance is achieved by connecting a second branch to \textit{feature} $\mathbf{ f}$, which we call domain classifier branch, as shown in Fig. \ref{fig:framework} and denoted as $G_{dom}$.

Different from the objectness predictor, the domain classifier receives regions from both set $\mathcal{S}$ and set $\mathcal{W}$ to predict the origin of the input regions ($\mathcal{S}$ or $\mathcal{W}$). It is also a binary classification task and the optimization objective used is similar to the objectness predictor $G_{obj}$:

\begin{equation}
\label{equ:dom}
L_{dom}{\left( \mathbf{w} \right)} = - \frac{1}{n} \sum_{i=1}^{n} [y_i^{dom} \log(p_i) + (1-y_i^{dom}) \log(1-p_i)]
\end{equation}
where $p_i = \frac{1}{1+\exp(-G_{dom}(\mathbf{f_i}))}$ is the probability that a region $\mathbf{r}_i$ belongs to set $\mathcal{S}$. In this domain classification task, the regions sampled from set $\mathcal{S}$ are positives, \ie, $y_i^{dom}=1$; the regions sampled from set $\mathcal{W}$ are negatives, \ie, $y_i^{dom}=0$.

During the forward propagation, the domain classifier proceeds standardly and calculates $L_{dom}$. While in the backward propagation process, the gradients from the domain classifier are reversed (multiplied by -1) before passed to $\mathbf{f}$. With this \textit{gradient reversal} operator, the network actually \textit{maximizes} $L_{dom}$ during the training process, which results in the \textit{incapability} of modeling the discriminative information between two domains, \ie, makes the internal representations $\mathbf{f}$ as indistinguishable as possible (\textit{domain-invariant}) for both domains. In our objectness model, domain invariance is achieved by learning objectness $G_{obj}(\mathbf{f})$ with such domain-invariant representations $\mathbf{f}$, which thus could be well transferred to target domains with unseen objects. \footnote{We refer the readers to the theoretical proof of the gradient reversal strategy in \cite{ganin2016domain}.}

At each training iteration, for objectness predictor, 64 annotated regions are sampled from set $\mathcal{S}$ to learn the objectness and the ratio of positives and negatives are balanced to 1:3. Meanwhile, for domain classifier, 64 unlabeled regions are randomly sampled from set $\mathcal{W}$. It should be noticed that directly utilizing 64 ``balanced'' regions of set $\mathcal{S}$ and 64 ``random'' regions from set $\mathcal{W}$ to achieve domain invariance is inappropriate, since the data distribution of set $\mathcal{S}$ has already been changed with the balancing process. To address this issue, we randomly sample \textit{another} 64 regions from set $\mathcal{S}$ to train the domain classifier together with the regions of set $\mathcal{W}$. Finally, each training mini-batch contains 192 samples, including 64 ``balanced'' regions sampled from set $\mathcal{S}$, another 64 ``random" regions sampled from set $\mathcal{S}$ and 64 ``random'' regions sampled from set $\mathcal{W}$. Note that the image-level labels are not used in this process. The detailed training strategy is described in Step 1 of Algorithm 1.

\begin{algorithm}
  \caption{Robust Objectness Transfer Approach for MSD}
  \begin{algorithmic}[l]
  \Require strong set $\mathcal{S}$ and its region proposals $\mathcal{R}_s$

  weak set $\mathcal{W}$ and its region proposals $\mathcal{R}_w$

  Objectness model $Obj$ with parameters $\mathcal{W}_{o}$

  Objectness-aware detection model $Det$ with parameters $\mathcal{W}_{d}$

  max iteration ${maxiter}_{obj}$ and ${maxiter}_{det}$ \State
  Step 1: Domain-invariant Objectness Learning \State
  Initialize $\mathcal{W}_{o}$
  \For{${iter}=1;$ ${iter}<{maxiter}_{obj};$ ++${iter}$} \State
  Sample 64 ``balanced'' regions from $\mathcal{R}_{s}$ \State
  Calculate objectness prediction loss using (\ref{equ:obj}) \State
  Propagate the \textit{gradients} \State
  Sample 64 ``random'' regions from $\mathcal{R}_{s}$ \State
  Sample 64 ``random'' regions from $\mathcal{R}_{w}$ \State
  Calculate domain classification loss using (\ref{equ:dom}) \State
  Propagate the \textit{\textbf{reversed} gradients} \State
  Update $\mathcal{W}_{o}$
  \EndFor\State
  Step 2: Objectness-aware Detection \State
  Initialize $\mathcal{W}_{d}$
  \For{${iter}=1;$ ${iter}<{maxiter}_{det};$ ++${iter}$} \State
  Sample one image $\mathbf{x}$ from $\mathcal{W}$ \State
  Scoring its region proposals $\mathbf{R}$ with $Obj$ \State
  Object bag $\mathbf{B}_{obj}$ $\gets$ the top 15\% regions of $\mathbf{R}$ \State
  Distractor bag $\mathbf{B}_{dis}$ $\gets$ the last 85\% regions of $\mathbf{R}$ \State
  Calculating the Loss using (\ref{equ:loss}) \State
  Update $\mathcal{W}_{d}$
  \EndFor\State
  \Ensure Detection model's parameter $\mathcal{W}_{d}$ \State
  \end{algorithmic}
\end{algorithm}

\subsection{Objectness-aware Detection Model}
\label{section:oaWSD}
After the objectness knowledge is learned, it is used to separate the objects and distractors in weakly labeled images of set $\mathcal{W}$. In our approach, for each weakly labeled image, its region proposals are first fed to the learned objectness model to get their ``objectness'' scores (\ie, the outputs of objectness predictor in Fig. \ref{fig:framework}), and then sorted according to the scores. The top m\% (m is set to 15 in our algorithm) proposals are selected as the ``object regions'', and the rest 1-m\% ones are used as ``distractor regions''. A simple way to utilize the selected object regions is to consider these regions as pseudo ground truth and then train fully supervised detectors. But it is not a well-performing approach since such separation between objects and distractors is not prominent. To address this issue, we aim to further model the difference between objects and distractors based on a multiple instance learning (MIL) approach and propose the objectness-aware detection model.

In MIL framework, we first construct the ``object bag'' with ``object regions'' and the ``distractor bag'' with ``distractor regions'' for each weakly labeled image in set $\mathcal{W}$. The labels for these bags are denoted as $y\in{\left\{ -1,1 \right\}}^{K+1}$. The ``distractor bags" are tagged with $y_0=1$ while the ``object bags" are labeled as their corresponding object categories ($y_k=1,k>0$).

Then we adopt the Fast R-CNN framework for the objectness-aware detection model. During training, a weakly labeled image $\mathbf{x}$ as well as its region proposals $R$ that are generated by selective search are imported as the input of the network (each image $\mathbf{x}$ contains two bags: the object bag and the distractor bag). The network simultaneously computes features for each proposal and finally maps the features to $K$+1-dimensional vectors ${s^R}\in{\mathbb{R}^{(K+1)\times{\left| R \right|}}}$, which represent the classification scores for regions. These region-level scores are directly used to evaluate the detection performance at testing time.

During training, regions in the bag cannot be labeled since we do not have bounding box annotations. Thus the region-level scores $s^R$ need to be aggregated to a bag-level classification score $s^B$ to train the model. In traditional MIL settings, the highest region-level score is selected as the bag-level score:
\begin{equation}
s^B = \underset{r}{\max}(s_{r}^{R}).
\end{equation}
This $\mathit{max}$ operator utilizes only one region per bag as the positive sample. To relax this restriction, we use ``$\mathit{exp}$-$\mathit{sum}$-$\mathit{log}$" operator proposed in \cite{bilen2014weakly} to serve as a soft approximation for the $\mathit{max}$ operator:
\begin{equation}
s^B=\log{\left( \sum_{r}^{\left| R \right|}{\exp{\left( s_r^R \right)}}\right)}.
\end{equation}

After obtaining the bag-level scores $s^B$, we utilize the sigmoid function to compute the posterior probability that each bag $\mathbf{B_i}$ belongs to the $k$-th class:
\begin{equation}
p_{\mathit{ki}} = \frac{1}{1+\exp(-s_{\mathit{ki}}^B)}.
\end{equation}

Finally, the network can be trained end-to-end using cross-entropy loss:
\begin{equation}
\label{equ:loss}
L{\left( \mathbf{w} \right)} = \frac{\lambda}{2}{\left\| \mathbf{w} \right\|}_2^2 \!- \frac{1}{n} \sum_{i=1}^{n} \sum_{k=1}^{K+1} \left( \mathbf{1}{\left\{ y_{\mathit{ki}}=1 \right\}} \log \left( {p_{\mathit{ki}}} \right) \right)
\end{equation}
where $y_i\in {\left\{ -1,1 \right\}}^{K+1}$ is the bag-level labels, and $\mathbf{1}{\left\{ \cdot \right\}}$ is the indicator function. $\lambda$ is the weight decay parameter on the weight $\mathbf{w}$ of CNNs used to improve the generalization of the model and is set to 0.0005 in all experiments. Using this MIL-based approach, the objectness-aware detection model is able to model the concepts of both distractors and $K$ object categories. Finally, with the learned \textit{distractor} concept, our method is able to reliably reject the distractors in images and significantly improve the detection performance on set $\mathcal{W}$. The overall approach is summarized in Algorithm 1.
\begin{table*}[t]
\renewcommand{\arraystretch}{1.3}
\caption{\label{tab:impl}\\The Summaries of Ours-MSD and Three Baselines.}
\vspace{-1em}
\begin{center}
\begin{tabular*}{\textwidth}{l @{\extracolsep{\fill}} c c}
\hline
         Method                              & Learning the knowledge from set $\mathcal{S}$ & Traning detectors on set $\mathcal{W}$       \\ \hline
Ours-MSD                                     & domain-invariant objectness model (VGG16)       & objectness-aware detection model (AlexNet/VGG16) \\
B-WSD                                        &  -                                            & standard MIL-based WSD model (AlexNet/VGG16) \\
B-MSD                                        & fully supervised Fast RCNN detector (VGG16)   & standard MIL-based WSD model (VGG16)         \\
OOM-MSD                                      & original objectness model (VGG16)             & objectness-aware detection model (AlexNet)   \\ \hline
\end{tabular*}
\end{center}
\footnotesize
\begin{tablenotes}
\item [1] \textit{The contents in parentheses indicate the architectures of the models, \ie, AlexNet or VGG16.}
    \vspace{-1em}
\end{tablenotes}
\end{table*}

\section{Experiments}

The proposed MSD method is evaluated on both intra-dataset detection task (Section \ref{section:intra}) and cross-dataset detection task (Section \ref{section:cross}). We use two metrics for evaluation: mAP and CorLoc. Mean Average Precision (mAP) is the evaluation metric to test our model on the test set, which follows the standard PASCAL VOC protocol \cite{everingham2005pascal}. Correct localization (CorLoc) is to test the proposed model on the training set measuring the localization accuracy \cite{deselaers2012weakly}. CorLoc is the percentage of images for which the most confident detected bounding box overlaps (IoU $\geq$ 0.5) with a ground truth box.

\subsection{Baselines}
\label{section:baseline}
In this section, we first introduce three baseline methods to compare with our robust objectness transfer MSD approach. For simplicity, we term our robust objectness transfer MSD approach as \textit{Ours-MSD} in the following sections.

\textit{B-WSD} (the \textbf{B}aseline \textbf{W}eakly \textbf{S}upervised \textbf{D}etection method). This baseline is a standard MIL-based WSD method, which is trained on set $\mathcal{W}$ only and does not utilize the objectness knowledge. Specially, in B-WSD, \textit{all} regions in an image construct an image bag tagged with its object category label. During training, B-WSD only aims to model the concepts of $K$ object categories and the loss function in (\ref{equ:loss}) is adapted to sum over $K$ categories.

\textit{B-MSD} (the \textbf{B}aseline \textbf{M}ixed \textbf{S}upervised \textbf{D}etection method). In Ours-MSD, the objectness knowledge is first modelled on set $\mathcal{S}$ and then transferred to set $\mathcal{W}$ to learn the objectness-aware detection model. In B-MSD, we utilize a straightforward \textit{fine-tuning} approach to transfer the knowledge of \textit{detection task} from set $\mathcal{S}$ to set $\mathcal{W}$. We first train a fully supervised Fast RCNN detector on set $\mathcal{S}$. Then we fine tune the obtained detector on set $\mathcal{W}$ under a MIL-based WSD framework. That is, we train a B-WSD model on set $\mathcal{W}$ initialized using the fully supervised detector.

\textit{OOM-MSD} (the \textbf{O}riginal \textbf{O}bjectness \textbf{M}odel for \textbf{M}ixed \textbf{S}upervised \textbf{D}etection). Similar to Ours-MSD, the OOM-MSD also utilizes objectness knowledge to separate the objects and distractors and then trains objectness-aware detection model. The difference lies in the architecture of the objectness model. In OOM-MSD, the domain adaptation component of the objectness model (\ie, the domain classifier branch in Fig. \ref{fig:framework}) is removed and the \textit{orginial} objectness knowledge is learned from set $\mathcal{S}$ only.

\subsection{Intra-dataset Detection}
\label{section:intra}

\subsubsection{Benchmark Data}
In the intra-dataset detection case, we evaluate our method on the PASCAL VOC 2007 \cite{everingham2005pascal} dataset and the ILSVRC2013 detection \cite{russakovsky2015imagenet} dataset.

The PASCAL VOC 2007 dataset contains 20 common object classes, 2,501 training images, 2,510 validation images and 5,011 test images. To train the objectness models and objectness-aware WSD models, we split the \textit{trainval} set (5,011 images in total) into two sets: images belonging to the first 10 categories of PASCAL constitute set $\mathcal{S}$ and images of the last 10 categories construct the set of weak categories (set $\mathcal{W}$). The strong set $\mathcal{S}$ includes 3,002 images. We have access to their bounding box annotations and model objectness knowledge from these annotated boxes. The weak set $\mathcal{W}$ contains 3,340 weakly labeled images, which are used to train objectness-aware detection models. The detection models are evaluated on \textit{test} set and the mAP is computed over the last 10 categories.

The ILSVRC2013 detection dataset contains 200 basic level object cateogories, 395,909 images for training, 20,121 images for validation, and 40,152 images for testing. The validation set is split in half: \textit{val1} and \textit{val2}, as in R-CNN \cite{girshick2014rich}. Then we collect images with bounding box annotations from both \textit{train} and \textit{val1} to construct our training set, \textit{trainval1} (107,452 images in total). Similar to PASCAL VOC, we also split the \textit{trainval1} set into two sets: the first 100 and the last 100 categories (in alphabetical order) correspond to the strong categories (54,735 images) and weak categories (57,584 images) respectively. Finally, the detection models are evaluated on \textit{val2} set (9,917 images) and the mAP is calculated over the last 100 categories.

It is noted that the training images in both datasets are possible to contain more than one object class. Thus a portion of training images would be included in both strong and weak sets. For example, an image containing both \textit{dog} (strong category) and \textit{person} (weak category) will be used in both sets. In this case, our method considers the image as a fully labeled \textit{dog} image in set $\mathcal{S}$ and a weakly labeled \textit{person} image in set $\mathcal{W}$.

\subsubsection{Implementation Details}
\label{section:intra-dataset-imple}
In this section, we introduce the detailed implementation settings for Ours-MSD and three baselines. Table \ref{tab:impl} summarizes their statistics.

\textit{Learning the Knowledge from Set $\mathcal{S}$.} In OOM-MSD baseline and Ours-MSD, we train the original objectness model and the domain-invariant objectness model from set $\mathcal{S}$ respectively. Both the two objectness models start from VGG16 models pre-trained on ImageNet classification \cite{simonyan2014very}. The last 1000-way fc layer of VGG16 is changed to a new 256-way fc layer and its output, the 256-dimensional vector, serves as the \textit{feature} $\mathbf{f}$ introduced in Section \ref{section:objectness}. The \textit{feature} $\mathbf{f}$ is then used for both objectness prediction and domain classification. For objectness prediction, $\mathbf{f}$ is directly connected to a 2-way fc layer; for domain classification, $\mathbf{f}$ is connected to 3 fc layers (1024-1024-2). Note that this domain classifier branch is removed in OOM-MSD.

During training, the objectness models are trained for 10 epochs using stochastic gradient descent (SGD). The initial learning rate is set to 0.001 for the first 5 epochs and decreased to 0.0001 for the last 5 epochs. A momentum 0.9 and a weights decay of 0.0005 are used. For domain-invariant model, the reversed gradients collected from domain classifier start from $-0.0\times\textit{gradient}$ and gradually decrease to $-0.1\times \textit{gradient}$ in the first 8,000 training iterations to make the training process stable in the early iterations.

In B-MSD baseline, we train a supervised Fast RCNN detector \cite{girshick2015fast} on set $\mathcal{S}$. The detector is initialized using VGG16 pre-trained on ImageNet classification, which is the same as the objectness models used in OOM-MSD and Ours-MSD. During training, the Fast RCNN detector is trained for 20 epochs, and the learning rates are set to 0.001 and 0.0001 for the first and the last 10 epochs respectively.

\textit{Training Object Detectors on Set $\mathcal{W}$.} For OOM-MSD and Ours-MSD, the objectness knowledge is utilized to train the objectness-aware detection models; for B-WSD and B-MSD, standard MIL-based WSD models are learned. As shown in Table \ref{tab:impl}, the implementations of the four detection models (Ours-MSD and three baselines) are different. The details are as follows:

In previous MSD method (\ie, LSSOD \cite{tang2016large}), the weakly supervised detectors are based on AlexNet \cite{krizhevsky2012imagenet}. Thus, for a fair comparison, the detection models used in B-WSD, OOM-MSD and Ours-MSD are initialized using the same AlexNet model pre-trained on ImageNet. The only exception is B-MSD. B-MSD proposes to fine tune VGG16 Fast RCNN detectors to the VGG16 WSD models. To compare with the B-MSD baseline, we additionally train detection models based on VGG16 for B-WSD baseline and Ours-MSD.

During training, the detection experiments run for 20 epochs and the learning rates are set to $5\times{10}^{-5}$ and $5\times{10}^{-6}$ respectively for the first and last 10 epochs. Similar to Fast RCNN \cite{girshick2015fast}, the aspect ratios of the input images are retained fixed, and the shorter sides of images are resized to 600. Only horizontal flipping is applied as a form of data augmentation. At testing time, non-maximum suppression (NMS) is used to ignore redundant, overlapped boxes and the threshold of NMS is set to 0.3.

\begin{table}[]
\renewcommand{\arraystretch}{1.1}
\captionsetup{justification=centering}
\caption{\label{tab:imagenet}\\Object Detection Performance (mAP \%) on ILSVRC2013 \textit{val2} Set.}
\vspace{-1em}
\begin{center}
{
\normalsize
\begin{tabular}{p{5cm}  p{2cm}<{\centering}}
\hline
         Method                              &\begin{tabular}[c]{@{}c@{}}mAP on set $\mathcal{W}$\\ (100 categories)\end{tabular}\\ \hline
weakly supervised: &  \\
B-WSD-AlexNet                                                                             & 13.78                                \\
B-WSD-VGG16                                                                                 & 11.82                               \\ \hline
(OM+MIL)+FRCN-AlexNet \cite{li2016weakly} $^*$                                              & 6.25                              \\
(OM+MIL)+FRCN-VGG16 \cite{li2016weakly} $^*$                                                & 9.10                               \\ \hline
\hline
mixed supervised: & \\
LSSOD (visual) \cite{tang2016large}            & 19.02                                \\
LSSOD (semantic) \cite{tang2016large}          & 19.04                                \\
ens-LSSOD \cite{tang2016large}                 & 20.03                                \\ \hline
B-MSD-VGG16                                      & 16.44                                   \\
OOM-MSD-AlexNet                                & 18.54                                \\
Ours-MSD-AlexNet                               & 22.28                                \\
Ours-MSD-VGG16                                   & \textbf{25.26}                       \\ \hline
\hline
fully supervised: & \\
FRCN-AlexNet: & 26.40 \\
FRCN-VGG16:   & 30.82 \\ \hline
\end{tabular}}
\end{center}
\footnotesize
\begin{tablenotes}
\item [1] \textit{The mAP is computed over the last 100 categories. $^*$ The results of \cite{li2016weakly} are obtained by averaging the reported numbers (in the supplementary material of \cite{li2016weakly}) over the last 100 categories.}
    \vspace{-1em}
\end{tablenotes}
\end{table}

\subsubsection{Evaluation on Large-scale Dataset}

\begin{figure*}[t]
\begin{center}
   \includegraphics[width=0.9\linewidth]{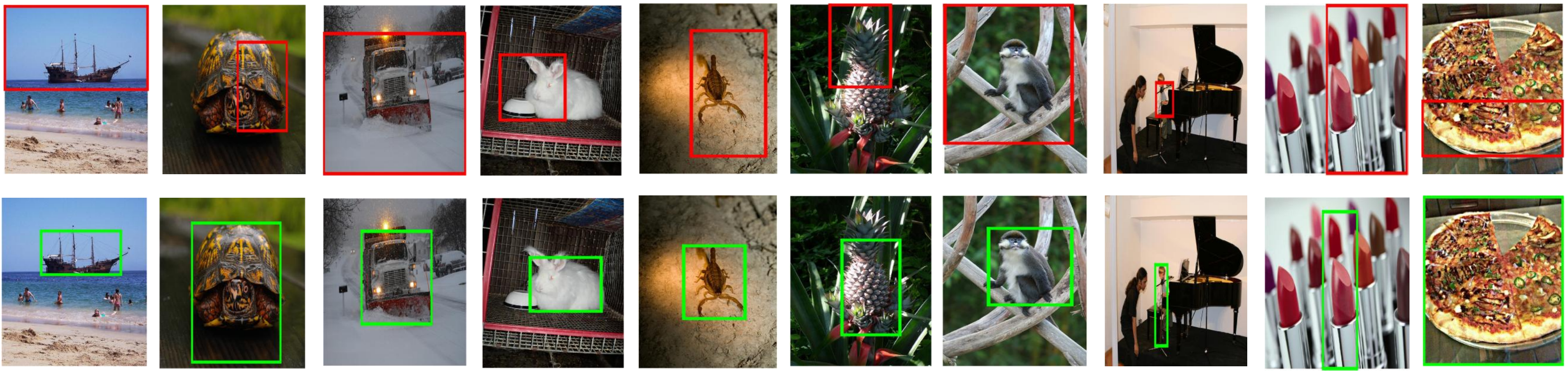}
\end{center}
   \caption{Sample detection results on ILSVRC2013 \textit{val2} set of the last 100 categories. The first row shows the results of the baseline WSD model (B-WSD-AlexNet) and the second row lists the results of the same images predicted by the objectness-aware model (Ours-MSD-AlexNet).}
\label{fig:results}
\end{figure*}

In this section, the proposed method is evaluated on ILSVRC2013 detection dataset. The experiment results (mAP \%) on set $\mathcal{W}$ (the last 100 categories of ILSVRC2013 detection) are shown in Table \ref{tab:imagenet}. We first focus on the WSD methods (the first compartment of Table \ref{tab:imagenet}). With the image-level labels only, B-WSD achieves a relatively low performance (B-WSD-AlexNet, 13.78 percent). When utilizing deeper networks, the performance of B-WSD is even lower (B-WSD-VGG16, 11.82 percent). The goal of B-WSD is to distinguish between different object categories. Thus, with more capable VGG16 networks, B-WSD might tend to search more discriminative object parts, rather than the whole objects. It will lead to inferior detection results. Regarding the B-WSD baseline, an interesting comparison is that the performance of B-WSD outperforms the state-of-the-art WSD result \cite{li2016weakly} by a large margin (13.78 versus 6.25 percent). While in PASCAL VOC dataset (as will be shown in Table \ref{tab:voc-cross-mAP}), \cite{li2016weakly} could easily surpass B-WSD (31.0 versus 23.87 percent). In the algorithm of \cite{li2016weakly}, several hyperparameters, \eg, the number of mined proposals, need to be set. In small dataset, such as PASCAL VOC, these hyperparameters could be decided by using cross-validation. However, in large-scale dataset, searching proper hyperparameters for hundreds of object categories is much more difficult and it finally results in unsatisfactory performance in large-scale scenarios. We believe that the robust algorithms with few hyperparameters are more appropriate, especially in large-scale datasets, to obtain high-performance detectors.

Different from the WSD methods, the OOM-MSD and Ours-MSD that are based on objectness transfer approach obtain much better results. Regarding the OOM-MSD baseline, as we have sufficient strong categories (the first 100 categories) in large-scale dataset, the original objectness model already learns, to some extent, ``general" objectness knowledge. Thus, the subsequent objectness-aware detection models could obtain remarkable improvement on set $\mathcal{W}$ (18.54 versus 13.78 percent). The performance of Ours-MSD is improved to 22.28 percent by applying domain-invariant model, where the learned objectness knowledge is more robust to the change of categories. Ours-MSD could obtain further improvements by utilizing deeper detection networks (Ours-MSD-VGG16, 25.26 percent). As the object parts are more likely recognized as \textit{distractors} and rejected with deeper networks in Ours-MSD. By fine tuning the supervised detectors to WSD models, the B-MSD baseline can improve the B-WSD baseline (16.44 versus 11.82 percent). But the performance is still obviously lower than Ours-MSD (16.44 versus 25.26 percent). In some way, with the fine-tuning approach, B-MSD aims to reduce the gap between the \textit{detection task} and the \textit{classification task}. In contrast, our MSD method transfers the \textit{objectness} knowledge to weak categories, which is crucial to detect objects. Moreover, our method also aims to reduce the discrepancy between different categories. The last compartment of Table \ref{tab:imagenet} shows the results of fully supervised Fast RCNN detectors trained on the last 100 categories. Compared with these oracle detectors, our MSD method attains comparable performance (\eg, 22.28 versus 26.40 percent).

In order to illustrate our improvements more clearly, we draw the detection results of the B-WSD and Ours-MSD for the same image, as shown in Fig. \ref{fig:results}. We can see that, in Ours-MSD, the objectness-aware model is endowed with the ability to reject the distractors (contexts or object parts) by incorporating the objectness knowledge. Such ability is essential for object detection and finally results in the remarkable improvements.

Our method is then compared with the state-of-the-art MSD method LSSOD \cite{tang2016large}, which performs intra-dataset detection in ILSVRC2013 (our experimental setup is slightly different from LSSOD, and we will clarify this in Appendix \ref{sec:appeimagenet}). LSSOD transfers the classifier and detector differences from strong to weak categories based on their visual and semantic similarities, which correspond to the visual transfer model and semantic transfer model respectively. The best performance of the two models are 19.02 (visual) and 19.04 percent (semantic). The results are slightly higher than the OOM-MSD baseline (18.54 percent), but they are lower than that of our method (22.28 percent). The ensemble of the two models (20.03 percent) still cannot compete with our method. These results demonstrate the effectiveness of our method on large-scale datasets and prove that making use of transferable objectness knowledge to improve WSD is reasonable and successful.

\begin{table}[]
\renewcommand{\arraystretch}{1.2}
\captionsetup{justification=centering}
\caption{\label{tab:voc-intra}\\Object Detection Performance (mAP \%) on PASCAL VOC 2007 \textit{test} Set.}
\normalsize
\begin{center}
\begin{tabular}{p{4.0cm}  p{2.0cm}<{\centering}}
\hline           Method           & \begin{tabular}[c]{@{}c@{}}mAP on set $\mathcal{W}$\\ (10 categories)\end{tabular} \\ \hline
B-WSD-AlexNet                       & 22.63                                        \\ \hline
                                                                              \hline
LSSOD (visual)              & 5.76                                         \\ \hline \hline
OOM-MSD-AlexNet                     & 14.00                                        \\
Ours-MSD-AlexNet                      & {\textbf{32.21}}                             \\ \hline
\end{tabular}
\end{center}
\footnotesize
\begin{tablenotes}
\item [1] \textit{The mAP is computed over the last 10 categories.}
    \vspace{-1em}
\end{tablenotes}
\end{table}

\begin{table*}[t]
\renewcommand{\arraystretch}{1.5}
\captionsetup{justification=centering}
\caption{\label{tab:voc-cross-mAP}\\Object Detection performance (mAP \%) on PASCAL VOC 2007 \textit{test} set.}
\vspace{-1em}
\begin{center}
\normalsize
\resizebox{\textwidth}{!}
{
\begin{tabular}{l llllllllllllllllllll l}
\hline
Method & \multicolumn{1}{c}{aero} & \multicolumn{1}{c}{bike} & \multicolumn{1}{c}{bird} & \multicolumn{1}{c}{boat} & \multicolumn{1}{c}{bottle} & \multicolumn{1}{c}{bus} &
\multicolumn{1}{c}{car} & \multicolumn{1}{c}{cat} & \multicolumn{1}{c}{chair} & \multicolumn{1}{c}{cow} & \multicolumn{1}{c}{table} & \multicolumn{1}{c}{dog} & \multicolumn{1}{c}{horse} & \multicolumn{1}{c}{mbike} & \multicolumn{1}{c}{person} & \multicolumn{1}{c}{plant} & \multicolumn{1}{c}{sheep} & \multicolumn{1}{c}{sofa} & \multicolumn{1}{c}{train} & \multicolumn{1}{c}{tv} & \multicolumn{1}{c}{mAP} \\ \hline
                                                                                  \hline
weakly supervised: &  &  &  &  &  &  &  &  &  &  &  &  &  &  &  &  &  &  &  & &  \\
B-WSD-AlexNet             & \multicolumn{1}{c}{40.5} & \multicolumn{1}{c}{35.3} & \multicolumn{1}{c}{19.5} & \multicolumn{1}{c}{5.8}  & \multicolumn{1}{c}{7.7}
                  & \multicolumn{1}{c}{38.9} & \multicolumn{1}{c}{39.9} & \multicolumn{1}{c}{23.3} & \multicolumn{1}{c}{1.6}  & \multicolumn{1}{c}{25.0}
                  & \multicolumn{1}{c}{11.1} & \multicolumn{1}{c}{25.2} & \multicolumn{1}{c}{29.9} & \multicolumn{1}{c}{49.5} & \multicolumn{1}{c}{21.3}
                  & \multicolumn{1}{c}{16.4} & \multicolumn{1}{c}{24.4} & \multicolumn{1}{c}{16.8} & \multicolumn{1}{c}{35.1} & \multicolumn{1}{c}{10.5}
                  & \multicolumn{1}{c}{23.87} \\
B-WSD-VGG16             & \multicolumn{1}{c}{43.1} & \multicolumn{1}{c}{32.5} & \multicolumn{1}{c}{18.5} & \multicolumn{1}{c}{8.7}  & \multicolumn{1}{c}{13.1}
                  & \multicolumn{1}{c}{33.5} & \multicolumn{1}{c}{37.3} & \multicolumn{1}{c}{18.0} & \multicolumn{1}{c}{11.4}  & \multicolumn{1}{c}{18.3}
                  & \multicolumn{1}{c}{21.8} & \multicolumn{1}{c}{28.5} & \multicolumn{1}{c}{23.3} & \multicolumn{1}{c}{46.4} & \multicolumn{1}{c}{8.5}
                  & \multicolumn{1}{c}{16.3} & \multicolumn{1}{c}{22.6} & \multicolumn{1}{c}{27.8} & \multicolumn{1}{c}{43.3} & \multicolumn{1}{c}{27.9}
                  & \multicolumn{1}{c}{25.03} \\ \hline
Wang \etal \cite{wang2014weakly}       & \multicolumn{1}{c}{48.9} & \multicolumn{1}{c}{42.3} & \multicolumn{1}{c}{26.1} & \multicolumn{1}{c}{11.3} & \multicolumn{1}{c}{11.9}
                  & \multicolumn{1}{c}{41.3} & \multicolumn{1}{c}{40.9} & \multicolumn{1}{c}{34.7} & \multicolumn{1}{c}{10.8} & \multicolumn{1}{c}{34.7}
                  & \multicolumn{1}{c}{18.8} & \multicolumn{1}{c}{34.4} & \multicolumn{1}{c}{35.4} & \multicolumn{1}{c}{52.7} & \multicolumn{1}{c}{19.1}
                  & \multicolumn{1}{c}{17.4} & \multicolumn{1}{c}{35.9} & \multicolumn{1}{c}{33.3} & \multicolumn{1}{c}{34.8} & \multicolumn{1}{c}{46.5}
                  & \multicolumn{1}{c}{31.6} \\ \hline
WSDDN-VGG$\_$F \cite{bilen2016weakly}   & \multicolumn{1}{c}{42.9} & \multicolumn{1}{c}{56.0} & \multicolumn{1}{c}{32.0} & \multicolumn{1}{c}{17.6} & \multicolumn{1}{c}{10.2}
                  & \multicolumn{1}{c}{61.8} & \multicolumn{1}{c}{50.2} & \multicolumn{1}{c}{29.0} & \multicolumn{1}{c}{3.8}  & \multicolumn{1}{c}{36.2}
                  & \multicolumn{1}{c}{18.5} & \multicolumn{1}{c}{31.1} & \multicolumn{1}{c}{45.8} & \multicolumn{1}{c}{54.5} & \multicolumn{1}{c}{10.2}
                  & \multicolumn{1}{c}{15.4} & \multicolumn{1}{c}{36.3} & \multicolumn{1}{c}{45.2} & \multicolumn{1}{c}{50.1} & \multicolumn{1}{c}{43.8}
                  & \multicolumn{1}{c}{34.5}  \\
WSDDN-VGG$\_$M \cite{bilen2016weakly}   & \multicolumn{1}{c}{43.6} & \multicolumn{1}{c}{50.4} & \multicolumn{1}{c}{32.2} & \multicolumn{1}{c}{26.0} & \multicolumn{1}{c}{9.8}
                  & \multicolumn{1}{c}{58.5} & \multicolumn{1}{c}{50.4} & \multicolumn{1}{c}{30.9} & \multicolumn{1}{c}{7.9}  & \multicolumn{1}{c}{36.1}
                  & \multicolumn{1}{c}{18.2} & \multicolumn{1}{c}{31.7} & \multicolumn{1}{c}{41.4} & \multicolumn{1}{c}{52.6} & \multicolumn{1}{c}{8.8}
                  & \multicolumn{1}{c}{14.0} & \multicolumn{1}{c}{37.8} & \multicolumn{1}{c}{46.9} & \multicolumn{1}{c}{53.4} & \multicolumn{1}{c}{47.9}
                  & \multicolumn{1}{c}{34.9}  \\
WSDDN-VGG16 \cite{bilen2016weakly}      & \multicolumn{1}{c}{39.4} & \multicolumn{1}{c}{50.1} & \multicolumn{1}{c}{31.5} & \multicolumn{1}{c}{16.3} & \multicolumn{1}{c}{12.6}
                  & \multicolumn{1}{c}{64.5} & \multicolumn{1}{c}{42.8} & \multicolumn{1}{c}{42.6} & \multicolumn{1}{c}{10.1} & \multicolumn{1}{c}{35.7}
                  & \multicolumn{1}{c}{24.9} & \multicolumn{1}{c}{38.2} & \multicolumn{1}{c}{34.4} & \multicolumn{1}{c}{55.6} & \multicolumn{1}{c}{9.4}
                  & \multicolumn{1}{c}{14.7} & \multicolumn{1}{c}{30.2} & \multicolumn{1}{c}{40.7} & \multicolumn{1}{c}{54.7} & \multicolumn{1}{c}{46.9}
                  & \multicolumn{1}{c}{34.8}  \\
WSDDN-Ens \cite{bilen2016weakly}        & \multicolumn{1}{c}{46.4} & \multicolumn{1}{c}{58.3} & \multicolumn{1}{c}{35.5} & \multicolumn{1}{c}{25.9} & \multicolumn{1}{c}{14.0}
                  & \multicolumn{1}{c}{66.7} & \multicolumn{1}{c}{53.0} & \multicolumn{1}{c}{39.2} & \multicolumn{1}{c}{8.9} & \multicolumn{1}{c}{41.8}
                  & \multicolumn{1}{c}{26.6} & \multicolumn{1}{c}{38.6} & \multicolumn{1}{c}{44.7} & \multicolumn{1}{c}{59.0} & \multicolumn{1}{c}{10.8}
                  & \multicolumn{1}{c}{17.3} & \multicolumn{1}{c}{40.7} & \multicolumn{1}{c}{49.6} & \multicolumn{1}{c}{56.9} & \multicolumn{1}{c}{50.8}
                  & \multicolumn{1}{c}{39.3} \\ \hline
(OM+MIL)+FRCN-AlexNet \cite{li2016weakly} & \multicolumn{1}{c}{49.7} & \multicolumn{1}{c}{33.6} & \multicolumn{1}{c}{30.8} & \multicolumn{1}{c}{19.9} & \multicolumn{1}{c}{13.0}
                  & \multicolumn{1}{c}{40.5} & \multicolumn{1}{c}{54.3} & \multicolumn{1}{c}{37.4} & \multicolumn{1}{c}{14.8} & \multicolumn{1}{c}{39.8}
                  & \multicolumn{1}{c}{9.4}  & \multicolumn{1}{c}{28.8} & \multicolumn{1}{c}{38.1} & \multicolumn{1}{c}{49.8} & \multicolumn{1}{c}{14.5}
                  & \multicolumn{1}{c}{24.0} & \multicolumn{1}{c}{27.1} & \multicolumn{1}{c}{12.1} & \multicolumn{1}{c}{42.3} & \multicolumn{1}{c}{39.7}
                  & \multicolumn{1}{c}{31.0} \\
(OM+MIL)+FRCN-VGG16 \cite{li2016weakly} & \multicolumn{1}{c}{54.5} & \multicolumn{1}{c}{47.4} & \multicolumn{1}{c}{41.3} & \multicolumn{1}{c}{20.8} & \multicolumn{1}{c}{17.7}
                  & \multicolumn{1}{c}{51.9} & \multicolumn{1}{c}{63.5} & \multicolumn{1}{c}{46.1} & \multicolumn{1}{c}{\textbf{21.8}} & \multicolumn{1}{c}{57.1}
                  & \multicolumn{1}{c}{22.1}  & \multicolumn{1}{c}{34.4} & \multicolumn{1}{c}{50.5} & \multicolumn{1}{c}{61.8} & \multicolumn{1}{c}{16.2}
                  & \multicolumn{1}{c}{29.9} & \multicolumn{1}{c}{40.7} & \multicolumn{1}{c}{15.9} & \multicolumn{1}{c}{55.3} & \multicolumn{1}{c}{40.2}
                  & \multicolumn{1}{c}{39.5} \\ \hline
ContextLocNet \cite{kantorov2016contextlocnet} & \multicolumn{1}{c}{57.1} & \multicolumn{1}{c}{52.0} & \multicolumn{1}{c}{31.5} & \multicolumn{1}{c}{7.6} & \multicolumn{1}{c}{11.5}
                  & \multicolumn{1}{c}{55.0} & \multicolumn{1}{c}{53.1} & \multicolumn{1}{c}{34.1} & \multicolumn{1}{c}{1.7} & \multicolumn{1}{c}{33.1}
                  & \multicolumn{1}{c}{\textbf{49.2}}  & \multicolumn{1}{c}{42.0} & \multicolumn{1}{c}{47.3} & \multicolumn{1}{c}{56.6} & \multicolumn{1}{c}{15.3}
                  & \multicolumn{1}{c}{12.8} & \multicolumn{1}{c}{24.8} & \multicolumn{1}{c}{48.9} & \multicolumn{1}{c}{44.4} & \multicolumn{1}{c}{47.8}
                  & \multicolumn{1}{c}{36.3} \\ \hline
WCCN-AlexNet \cite{diba2016weakly} & \multicolumn{1}{c}{43.9} & \multicolumn{1}{c}{57.6} & \multicolumn{1}{c}{34.9} & \multicolumn{1}{c}{21.3} & \multicolumn{1}{c}{14.7}
                  & \multicolumn{1}{c}{64.7} & \multicolumn{1}{c}{52.8} & \multicolumn{1}{c}{34.2} & \multicolumn{1}{c}{6.5} & \multicolumn{1}{c}{41.2}
                  & \multicolumn{1}{c}{20.5}  & \multicolumn{1}{c}{33.8} & \multicolumn{1}{c}{47.6} & \multicolumn{1}{c}{56.8} & \multicolumn{1}{c}{12.7}
                  & \multicolumn{1}{c}{18.8} & \multicolumn{1}{c}{39.6} & \multicolumn{1}{c}{46.9} & \multicolumn{1}{c}{52.9} & \multicolumn{1}{c}{45.1}
                  & \multicolumn{1}{c}{37.3} \\
WCCN-VGG16 \cite{diba2016weakly} & \multicolumn{1}{c}{49.5} & \multicolumn{1}{c}{60.6} & \multicolumn{1}{c}{38.6} & \multicolumn{1}{c}{29.2} & \multicolumn{1}{c}{16.2}
                  & \multicolumn{1}{c}{\textbf{70.8}} & \multicolumn{1}{c}{56.9} & \multicolumn{1}{c}{42.5} & \multicolumn{1}{c}{10.9} & \multicolumn{1}{c}{44.1}
                  & \multicolumn{1}{c}{29.9}  & \multicolumn{1}{c}{42.2} & \multicolumn{1}{c}{47.9} & \multicolumn{1}{c}{64.1} & \multicolumn{1}{c}{13.8}
                  & \multicolumn{1}{c}{23.5} & \multicolumn{1}{c}{45.9} & \multicolumn{1}{c}{54.1} & \multicolumn{1}{c}{\textbf{60.8}} & \multicolumn{1}{c}{54.5}
                  & \multicolumn{1}{c}{42.8} \\ \hline
                                                 \hline
mixed supervised: &  &  &  &  &  &  &  &  &  &  &  &  &  &  &  &  &  &  &  & &  \\
B-MSD-VGG16       & \multicolumn{1}{c}{40.1} & \multicolumn{1}{c}{49.2} & \multicolumn{1}{c}{31.2} & \multicolumn{1}{c}{13.9} & \multicolumn{1}{c}{23.9}
                  & \multicolumn{1}{c}{48.7} & \multicolumn{1}{c}{52.9} & \multicolumn{1}{c}{42.6} & \multicolumn{1}{c}{4.9} & \multicolumn{1}{c}{40.5}
                  & \multicolumn{1}{c}{26.1} & \multicolumn{1}{c}{38.5} & \multicolumn{1}{c}{36.5} & \multicolumn{1}{c}{61.3} & \multicolumn{1}{c}{\textbf{22.3}}
                  & \multicolumn{1}{c}{14.6} & \multicolumn{1}{c}{39.1} & \multicolumn{1}{c}{22.4} & \multicolumn{1}{c}{37.4} & \multicolumn{1}{c}{24.7}
                  & \multicolumn{1}{c}{33.53}  \\
OOM-MSD-AlexNet   & \multicolumn{1}{c}{52.5} & \multicolumn{1}{c}{49.1} & \multicolumn{1}{c}{37.9} & \multicolumn{1}{c}{35.1} & \multicolumn{1}{c}{15.5}
                  & \multicolumn{1}{c}{49.9} & \multicolumn{1}{c}{51.9} & \multicolumn{1}{c}{53.2} & \multicolumn{1}{c}{8.9} & \multicolumn{1}{c}{40.1}
                  & \multicolumn{1}{c}{17.9} & \multicolumn{1}{c}{47.6} & \multicolumn{1}{c}{50.9} & \multicolumn{1}{c}{59.3} & \multicolumn{1}{c}{11.6}
                  & \multicolumn{1}{c}{16.8} & \multicolumn{1}{c}{43.7} & \multicolumn{1}{c}{26.3} & \multicolumn{1}{c}{43.6} & \multicolumn{1}{c}{40.9}
                  & \multicolumn{1}{c}{37.65}  \\
Ours-MSD-AlexNet  & \multicolumn{1}{c}{55.8} & \multicolumn{1}{c}{56.6} & \multicolumn{1}{c}{41.1} & \multicolumn{1}{c}{35.1} & \multicolumn{1}{c}{22.8}
                  & \multicolumn{1}{c}{60.1} & \multicolumn{1}{c}{58.5} & \multicolumn{1}{c}{55.0} & \multicolumn{1}{c}{10.3} & \multicolumn{1}{c}{48.5}
                  & \multicolumn{1}{c}{22.2} & \multicolumn{1}{c}{50.5} & \multicolumn{1}{c}{55.8} & \multicolumn{1}{c}{61.6} & \multicolumn{1}{c}{12.8}
                  & \multicolumn{1}{c}{21.7} & \multicolumn{1}{c}{44.4} & \multicolumn{1}{c}{26.1} & \multicolumn{1}{c}{46.8} & \multicolumn{1}{c}{49.4}
                  & \multicolumn{1}{c}{41.77}  \\
Ours-MSD-VGG16    & \multicolumn{1}{c}{61.1} & \multicolumn{1}{c}{66.3} & \multicolumn{1}{c}{\textbf{54.8}} & \multicolumn{1}{c}{12.5} & \multicolumn{1}{c}{\textbf{36.5}}
                  & \multicolumn{1}{c}{63.1} & \multicolumn{1}{c}{61.9} & \multicolumn{1}{c}{\textbf{66.9}} & \multicolumn{1}{c}{17.5} & \multicolumn{1}{c}{66.1}
                  & \multicolumn{1}{c}{14.3} & \multicolumn{1}{c}{69.3} & \multicolumn{1}{c}{65.4} & \multicolumn{1}{c}{69.6} & \multicolumn{1}{c}{2.4}
                  & \multicolumn{1}{c}{20.5} & \multicolumn{1}{c}{54.6} & \multicolumn{1}{c}{34.3} & \multicolumn{1}{c}{58.3} & \multicolumn{1}{c}{54.6}
                  & \multicolumn{1}{c}{47.50}  \\
Ours-MSD-Ens      & \multicolumn{1}{c}{65.6} & \multicolumn{1}{c}{67.2} & \multicolumn{1}{c}{54.3} & \multicolumn{1}{c}{31.2} & \multicolumn{1}{c}{35.8}
                  & \multicolumn{1}{c}{68.1} & \multicolumn{1}{c}{65.0} & \multicolumn{1}{c}{63.3} & \multicolumn{1}{c}{17.3} & \multicolumn{1}{c}{\textbf{66.8}}
                  & \multicolumn{1}{c}{18.3} & \multicolumn{1}{c}{70.6} & \multicolumn{1}{c}{66.7} & \multicolumn{1}{c}{\textbf{69.8}} & \multicolumn{1}{c}{3.7}
                  & \multicolumn{1}{c}{24.7} & \multicolumn{1}{c}{\textbf{55.0}} & \multicolumn{1}{c}{37.4} & \multicolumn{1}{c}{58.3} & \multicolumn{1}{c}{57.3}
                  & \multicolumn{1}{c}{49.82}  \\
Ours-MSD-Ens + FRCN-VGG16    & \multicolumn{1}{c}{\textbf{70.5}} & \multicolumn{1}{c}{\textbf{69.2}} & \multicolumn{1}{c}{53.3} & \multicolumn{1}{c}{\textbf{43.7}} & \multicolumn{1}{c}{25.4}
                  & \multicolumn{1}{c}{68.9} & \multicolumn{1}{c}{\textbf{68.7}} & \multicolumn{1}{c}{56.9} & \multicolumn{1}{c}{18.4} & \multicolumn{1}{c}{64.2}
                  & \multicolumn{1}{c}{15.3} & \multicolumn{1}{c}{\textbf{72.0}} & \multicolumn{1}{c}{\textbf{74.4}} & \multicolumn{1}{c}{65.2} & \multicolumn{1}{c}{15.4}
                  & \multicolumn{1}{c}{\textbf{25.1}} & \multicolumn{1}{c}{53.6} & \multicolumn{1}{c}{\textbf{54.4}} & \multicolumn{1}{c}{45.6} & \multicolumn{1}{c}{\textbf{61.4}}
                  & \multicolumn{1}{c}{\textbf{51.08}}  \\ \hline
\hline
fully supervised: &  &  &  &  &  &  &  &  &  &  &  &  &  &  &  &  &  &  &  & &  \\
FRCN-AlexNet      & \multicolumn{1}{c}{61.5} & \multicolumn{1}{c}{64.6} & \multicolumn{1}{c}{46.3} & \multicolumn{1}{c}{34.3} & \multicolumn{1}{c}{20.4}
                  & \multicolumn{1}{c}{65.0} & \multicolumn{1}{c}{65.9} & \multicolumn{1}{c}{63.8} & \multicolumn{1}{c}{27.0} & \multicolumn{1}{c}{57.2}
                  & \multicolumn{1}{c}{55.0} & \multicolumn{1}{c}{54.3} & \multicolumn{1}{c}{62.0} & \multicolumn{1}{c}{66.8} & \multicolumn{1}{c}{48.0}
                  & \multicolumn{1}{c}{23.7} & \multicolumn{1}{c}{49.0} & \multicolumn{1}{c}{49.0} & \multicolumn{1}{c}{62.7} & \multicolumn{1}{c}{58.6}
                  & \multicolumn{1}{c}{51.8}  \\
FRCN-VGG16      & \multicolumn{1}{c}{71.7} & \multicolumn{1}{c}{72.0} & \multicolumn{1}{c}{60.8} & \multicolumn{1}{c}{45.1} & \multicolumn{1}{c}{32.3}
                  & \multicolumn{1}{c}{73.6} & \multicolumn{1}{c}{76.6} & \multicolumn{1}{c}{78.7} & \multicolumn{1}{c}{35.8} & \multicolumn{1}{c}{72.3}
                  & \multicolumn{1}{c}{62.8} & \multicolumn{1}{c}{75.1} & \multicolumn{1}{c}{71.1} & \multicolumn{1}{c}{73.0} & \multicolumn{1}{c}{61.2}
                  & \multicolumn{1}{c}{32.0} & \multicolumn{1}{c}{62.3} & \multicolumn{1}{c}{65.6} & \multicolumn{1}{c}{70.6} & \multicolumn{1}{c}{65.5}
                  & \multicolumn{1}{c}{62.9}  \\ \hline
\end{tabular}
}
\footnotesize
\begin{tablenotes}
\item [1] \textit{``Ours-MSD-Ens'' is the ensemble model obtained by averaging the outputs of AlexNet and VGG16 objectness-aware models. ``Ours-MSD-Ens + FRCN-VGG16'' indicates the method that utilizes our ensemble model to select top-scoring regions as pseudo ground truths and then trains a supervised Fast RCNN detector \cite{girshick2015fast} using VGG16 model.}
    \vspace{-1em}
\end{tablenotes}
\end{center}
\end{table*}

\subsubsection{Evaluation on Smaller Dataset}
We also compare these methods on a much smaller dataset, PASCAL VOC 2007, where only a small quantity of fully labeled categories are available. The results are shown in Table \ref{tab:voc-intra}. In the smaller dataset, the distribution discrepancy between strong and weak categories becomes very large. The original objectness model trained with only 10 strong categories will have strong bias on these categories, and the subsequent objectness-aware detection model only achieves 14.00 percent on weak categories, which is even far below the baseline method (22.63 percent). LSSOD cannot deal with the large distribution discrepancy as well. During the knowledge transfer process, LSSOD assumes that the differences learned on layers 1-7 in AlexNet are category-\textit{invariant}, and those differences are shared between strong and weak categories. However, with large distribution discrepancy, layers 1-7 actually learn category-\textit{specific} representations, and this assumption is no longer hold. As a result, LSSOD completely fails (5.76 percent) in this scenario.

On the contrary, we can relieve these impacts by training domain-invariant objectness model. With the domain-invariant knowledge, our method significantly improves the detection performance to 32.21 percent. It proves the ability of our domain-invariant method to cope with the large distribution discrepancy. Even if we only have access to limited fully labeled categories, we still be able to use the proposed method to localize new objects.

\subsection{Cross-dataset Detection}
\label{section:cross}

\subsubsection{Benchmark Data}
In this section, we evaluate our MSD framework on a cross-dataset detection task. The task is conducted between PASCAL VOC 2007 dataset and ILSVRC2013 detection dataset. In the cross-dataset scenario, all the images in VOC 2007 \textit{trainval} constitute set $\mathcal{W}$ (20 categories, 5,011 images); the images whose categories do not overlap with set $\mathcal{W}$ are selected from the ILSVRC2013 \textit{trainval1} to form the set $\mathcal{S}$ (180 categories, 89,391 images). The mAP is used to evaluate the performance of the detection models on VOC 2007 \textit{test} set over all the 20 categories. The CorLoc \cite{deselaers2012weakly} is applied to measure the localization accuracy of models on VOC 2007 \textit{trainval} set. We also conduct the cross-dataset detection on PASCAL VOC 2010 and PASCAL VOC 2012. The object categories of VOC 2010 and VOC 2012 are the same as VOC 2007.
The size of VOC 2010 and VOC 2012 is approximately twice larger than VOC 2007 for both \textit{trainval} and \textit{test} sets. For VOC 2010, the \textit{trainval} images construct set $\mathcal{W}$ and the detection performance is evaluated on VOC 2010 \textit{test} set over all the 20 categories. The same cross-dataset setting is adopted for VOC 2012 dataset.

\begin{table*}[]
\renewcommand{\arraystretch}{1.5}
\captionsetup{justification=centering}
\caption{\label{tab:voc-cross-corloc}\\Object Detection Performance (CorLoc \%) on PASCAL VOC 2007 \textit{trainval} Set}
\vspace{-1em}
\begin{center}
\normalsize
\resizebox{\textwidth}{!} {
\begin{tabular}{l llllllllllllllllllll l }
\hline
Method & \multicolumn{1}{c}{aero} & \multicolumn{1}{c}{bike} & \multicolumn{1}{c}{bird} & \multicolumn{1}{c}{boat} & \multicolumn{1}{c}{bottle} & \multicolumn{1}{c}{bus} &
\multicolumn{1}{c}{car} & \multicolumn{1}{c}{cat} & \multicolumn{1}{c}{chair} & \multicolumn{1}{c}{cow} & \multicolumn{1}{c}{table} & \multicolumn{1}{c}{dog} & \multicolumn{1}{c}{horse} & \multicolumn{1}{c}{mbike} & \multicolumn{1}{c}{person} & \multicolumn{1}{c}{plant} & \multicolumn{1}{c}{sheep} & \multicolumn{1}{c}{sofa} & \multicolumn{1}{c}{train} & \multicolumn{1}{c}{tv} & \multicolumn{1}{c}{CorLoc} \\ \hline
                                                                                  \hline
weakly supervised: &  &  &  &  &  &  &  &  &  &  &  &  &  &  &  &  &  &  &  & &  \\
B-WSD-AlexNet             & \multicolumn{1}{c}{67.50} & \multicolumn{1}{c}{54.51} & \multicolumn{1}{c}{36.04} & \multicolumn{1}{c}{20.21}  & \multicolumn{1}{c}{14.12}
                  & \multicolumn{1}{c}{54.31} & \multicolumn{1}{c}{66.10} & \multicolumn{1}{c}{49.13} & \multicolumn{1}{c}{9.27}  & \multicolumn{1}{c}{54.79}
                  & \multicolumn{1}{c}{11.79} & \multicolumn{1}{c}{41.63} & \multicolumn{1}{c}{45.58} & \multicolumn{1}{c}{75.10} & \multicolumn{1}{c}{38.04}
                  & \multicolumn{1}{c}{40.00} & \multicolumn{1}{c}{48.45} & \multicolumn{1}{c}{30.65} & \multicolumn{1}{c}{50.95} & \multicolumn{1}{c}{21.86}
                  & \multicolumn{1}{c}{41.35} \\
B-WSD-VGG16             & \multicolumn{1}{c}{76.67} & \multicolumn{1}{c}{58.82} & \multicolumn{1}{c}{39.04} & \multicolumn{1}{c}{26.06}  & \multicolumn{1}{c}{34.73}
                  & \multicolumn{1}{c}{58.88} & \multicolumn{1}{c}{72.40} & \multicolumn{1}{c}{48.55} & \multicolumn{1}{c}{26.57}  & \multicolumn{1}{c}{43.15}
                  & \multicolumn{1}{c}{42.21} & \multicolumn{1}{c}{43.95} & \multicolumn{1}{c}{48.98} & \multicolumn{1}{c}{78.71} & \multicolumn{1}{c}{28.07}
                  & \multicolumn{1}{c}{38.46} & \multicolumn{1}{c}{50.52} & \multicolumn{1}{c}{43.01} & \multicolumn{1}{c}{59.32} & \multicolumn{1}{c}{41.94}
                  & \multicolumn{1}{c}{48.00} \\ \hline
Wang \etal \cite{wang2014weakly}        & \multicolumn{1}{c}{80.1} & \multicolumn{1}{c}{63.9} & \multicolumn{1}{c}{51.5} & \multicolumn{1}{c}{14.9} & \multicolumn{1}{c}{21.0}
                  & \multicolumn{1}{c}{55.7} & \multicolumn{1}{c}{74.2} & \multicolumn{1}{c}{43.5} & \multicolumn{1}{c}{26.2} & \multicolumn{1}{c}{53.4}
                  & \multicolumn{1}{c}{16.3} & \multicolumn{1}{c}{56.7} & \multicolumn{1}{c}{58.3} & \multicolumn{1}{c}{69.5} & \multicolumn{1}{c}{14.1}
                  & \multicolumn{1}{c}{38.3} & \multicolumn{1}{c}{58.8} & \multicolumn{1}{c}{47.2} & \multicolumn{1}{c}{49.1} & \multicolumn{1}{c}{60.9}
                  & \multicolumn{1}{c}{48.5} \\ \hline
WSDDN-VGG$\_$F \cite{bilen2016weakly}   & \multicolumn{1}{c}{68.5} & \multicolumn{1}{c}{67.5} & \multicolumn{1}{c}{56.7} & \multicolumn{1}{c}{34.3} & \multicolumn{1}{c}{32.8}
                  & \multicolumn{1}{c}{69.9} & \multicolumn{1}{c}{75.0} & \multicolumn{1}{c}{45.7} & \multicolumn{1}{c}{17.1}  & \multicolumn{1}{c}{68.1}
                  & \multicolumn{1}{c}{30.5} & \multicolumn{1}{c}{40.6} & \multicolumn{1}{c}{67.2} & \multicolumn{1}{c}{82.9} & \multicolumn{1}{c}{28.8}
                  & \multicolumn{1}{c}{43.7} & \multicolumn{1}{c}{71.9} & \multicolumn{1}{c}{62.0} & \multicolumn{1}{c}{62.8} & \multicolumn{1}{c}{58.2}
                  & \multicolumn{1}{c}{54.2}  \\
WSDDN-VGG$\_$M \cite{bilen2016weakly}   & \multicolumn{1}{c}{65.1} & \multicolumn{1}{c}{63.4} & \multicolumn{1}{c}{59.7} & \multicolumn{1}{c}{45.9} & \multicolumn{1}{c}{38.5}
                  & \multicolumn{1}{c}{69.4} & \multicolumn{1}{c}{77.0} & \multicolumn{1}{c}{50.7} & \multicolumn{1}{c}{30.1}  & \multicolumn{1}{c}{68.8}
                  & \multicolumn{1}{c}{34.0} & \multicolumn{1}{c}{37.3} & \multicolumn{1}{c}{61.0} & \multicolumn{1}{c}{82.9} & \multicolumn{1}{c}{25.1}
                  & \multicolumn{1}{c}{42.9} & \multicolumn{1}{c}{79.2} & \multicolumn{1}{c}{59.4} & \multicolumn{1}{c}{68.2} & \multicolumn{1}{c}{64.1}
                  & \multicolumn{1}{c}{56.1}  \\
WSDDN-VGG16 \cite{bilen2016weakly}      & \multicolumn{1}{c}{65.1} & \multicolumn{1}{c}{58.8} & \multicolumn{1}{c}{58.5} & \multicolumn{1}{c}{33.1} & \multicolumn{1}{c}{39.8}
                  & \multicolumn{1}{c}{68.3} & \multicolumn{1}{c}{60.2} & \multicolumn{1}{c}{59.6} & \multicolumn{1}{c}{\textbf{34.8}} & \multicolumn{1}{c}{64.5}
                  & \multicolumn{1}{c}{\textbf{60.5}} & \multicolumn{1}{c}{43.0} & \multicolumn{1}{c}{56.8} & \multicolumn{1}{c}{82.4} & \multicolumn{1}{c}{25.5}
                  & \multicolumn{1}{c}{41.6} & \multicolumn{1}{c}{61.5} & \multicolumn{1}{c}{55.9} & \multicolumn{1}{c}{65.9} & \multicolumn{1}{c}{63.7}
                  & \multicolumn{1}{c}{53.5}  \\
WSDDN-Ens \cite{bilen2016weakly}        & \multicolumn{1}{c}{68.9} & \multicolumn{1}{c}{68.7} & \multicolumn{1}{c}{65.2} & \multicolumn{1}{c}{42.5} & \multicolumn{1}{c}{40.6}
                  & \multicolumn{1}{c}{72.6} & \multicolumn{1}{c}{75.2} & \multicolumn{1}{c}{53.7} & \multicolumn{1}{c}{29.7} & \multicolumn{1}{c}{68.1}
                  & \multicolumn{1}{c}{33.5} & \multicolumn{1}{c}{45.6} & \multicolumn{1}{c}{65.9} & \multicolumn{1}{c}{86.1} & \multicolumn{1}{c}{27.5}
                  & \multicolumn{1}{c}{44.9} & \multicolumn{1}{c}{76.0} & \multicolumn{1}{c}{62.4} & \multicolumn{1}{c}{66.3} & \multicolumn{1}{c}{66.8}
                  & \multicolumn{1}{c}{58.0} \\ \hline
(OM+MIL)+FRCN-AlexNet \cite{li2016weakly} & \multicolumn{1}{c}{77.3} & \multicolumn{1}{c}{62.6} & \multicolumn{1}{c}{53.3} & \multicolumn{1}{c}{41.4} & \multicolumn{1}{c}{28.7}
                  & \multicolumn{1}{c}{58.6} & \multicolumn{1}{c}{76.2} & \multicolumn{1}{c}{61.1} & \multicolumn{1}{c}{24.5} & \multicolumn{1}{c}{59.6}
                  & \multicolumn{1}{c}{18.0}  & \multicolumn{1}{c}{49.9} & \multicolumn{1}{c}{56.8} & \multicolumn{1}{c}{71.4} & \multicolumn{1}{c}{20.9}
                  & \multicolumn{1}{c}{44.5} & \multicolumn{1}{c}{59.4} & \multicolumn{1}{c}{22.3} & \multicolumn{1}{c}{60.9} & \multicolumn{1}{c}{48.8}
                  & \multicolumn{1}{c}{49.8} \\
(OM+MIL)+FRCN-VGG16 \cite{li2016weakly} & \multicolumn{1}{c}{78.2} &\multicolumn{1}{c}{67.1} & \multicolumn{1}{c}{61.8} & \multicolumn{1}{c}{38.1} & \multicolumn{1}{c}{36.1}
                  & \multicolumn{1}{c}{61.8} & \multicolumn{1}{c}{78.8} & \multicolumn{1}{c}{55.2} & \multicolumn{1}{c}{28.5} & \multicolumn{1}{c}{68.8}
                  & \multicolumn{1}{c}{18.5}  & \multicolumn{1}{c}{49.2} & \multicolumn{1}{c}{64.1} & \multicolumn{1}{c}{73.5} & \multicolumn{1}{c}{21.4}
                  & \multicolumn{1}{c}{47.4} & \multicolumn{1}{c}{64.6} & \multicolumn{1}{c}{22.3} & \multicolumn{1}{c}{60.9} & \multicolumn{1}{c}{52.3}
                  & \multicolumn{1}{c}{52.4} \\ \hline
ContextLocNet \cite{kantorov2016contextlocnet} & \multicolumn{1}{c}{83.3} & \multicolumn{1}{c}{68.6} & \multicolumn{1}{c}{54.7} & \multicolumn{1}{c}{23.4} & \multicolumn{1}{c}{18.3}
                  & \multicolumn{1}{c}{73.6} & \multicolumn{1}{c}{74.1} & \multicolumn{1}{c}{54.1} & \multicolumn{1}{c}{8.6} & \multicolumn{1}{c}{65.1}
                  & \multicolumn{1}{c}{47.1}  & \multicolumn{1}{c}{59.5} & \multicolumn{1}{c}{67.0} & \multicolumn{1}{c}{83.5} & \multicolumn{1}{c}{35.3}
                  & \multicolumn{1}{c}{39.9} & \multicolumn{1}{c}{67.0} & \multicolumn{1}{c}{49.7} & \multicolumn{1}{c}{63.5} & \multicolumn{1}{c}{65.2}
                  & \multicolumn{1}{c}{55.1} \\ \hline
WCCN-AlexNet \cite{diba2016weakly} & \multicolumn{1}{c}{79.7} & \multicolumn{1}{c}{68.1} & \multicolumn{1}{c}{60.4} & \multicolumn{1}{c}{38.9} & \multicolumn{1}{c}{36.8}
                  & \multicolumn{1}{c}{61.1} & \multicolumn{1}{c}{78.6} & \multicolumn{1}{c}{56.7} & \multicolumn{1}{c}{27.8} & \multicolumn{1}{c}{67.7}
                  & \multicolumn{1}{c}{20.3}  & \multicolumn{1}{c}{48.1} & \multicolumn{1}{c}{63.9} & \multicolumn{1}{c}{75.1} & \multicolumn{1}{c}{21.5}
                  & \multicolumn{1}{c}{46.9} & \multicolumn{1}{c}{64.8} & \multicolumn{1}{c}{23.4} & \multicolumn{1}{c}{60.2} & \multicolumn{1}{c}{52.4}
                  & \multicolumn{1}{c}{52.6} \\
WCCN-VGG16 \cite{diba2016weakly} & \multicolumn{1}{c}{83.9} & \multicolumn{1}{c}{72.8} & \multicolumn{1}{c}{64.5} & \multicolumn{1}{c}{44.1} & \multicolumn{1}{c}{40.1}
                  & \multicolumn{1}{c}{65.7} & \multicolumn{1}{c}{82.5} & \multicolumn{1}{c}{58.9} & \multicolumn{1}{c}{33.7} & \multicolumn{1}{c}{72.5}
                  & \multicolumn{1}{c}{25.6}  & \multicolumn{1}{c}{53.7} & \multicolumn{1}{c}{67.4} & \multicolumn{1}{c}{77.4} & \multicolumn{1}{c}{26.8}
                  & \multicolumn{1}{c}{\textbf{49.1}} & \multicolumn{1}{c}{68.1} & \multicolumn{1}{c}{27.9} & \multicolumn{1}{c}{64.5} & \multicolumn{1}{c}{55.7}
                  & \multicolumn{1}{c}{56.7} \\ \hline
WSLAT$_{\mathit{weak}}$ \cite{rochan2015weakly}       & \multicolumn{1}{c}{77.31} & \multicolumn{1}{c}{55.55} & \multicolumn{1}{c}{62.76} & \multicolumn{1}{c}{40.88} & \multicolumn{1}{c}{21.31}
                  & \multicolumn{1}{c}{77.96} & \multicolumn{1}{c}{72.1} & \multicolumn{1}{c}{54.9} & \multicolumn{1}{c}{14.83} & \multicolumn{1}{c}{68.79}
                  & \multicolumn{1}{c}{29.50} & \multicolumn{1}{c}{56.29} & \multicolumn{1}{c}{70.38} & \multicolumn{1}{c}{74.69} & \multicolumn{1}{c}{43.18}
                  & \multicolumn{1}{c}{27.35} & \multicolumn{1}{c}{47.91} & \multicolumn{1}{c}{26.20} & \multicolumn{1}{c}{70.88} & \multicolumn{1}{c}{67.19}
                  & \multicolumn{1}{c}{53.00} \\ \hline
                                                 \hline
mixed supervised: &  &  &  &  &  &  &  &  &  &  &  &  &  &  &  &  &  &  &  & &  \\
WSLAT$_{\mathit{trans}}$ \cite{rochan2015weakly}      & \multicolumn{1}{c}{48.32} & \multicolumn{1}{c}{48.97} & \multicolumn{1}{c}{17.58} & \multicolumn{1}{c}{55.25} & \multicolumn{1}{c}{6.15}
                  & \multicolumn{1}{c}{32.26} & \multicolumn{1}{c}{15.85} & \multicolumn{1}{c}{40.36} & \multicolumn{1}{c}{28.54} & \multicolumn{1}{c}{70.92}
                  & \multicolumn{1}{c}{4.50} & \multicolumn{1}{c}{15.91} & \multicolumn{1}{c}{43.55} & \multicolumn{1}{c}{34.69} & \multicolumn{1}{c}{13.75}
                  & \multicolumn{1}{c}{3.26} & \multicolumn{1}{c}{51.04} & \multicolumn{1}{c}{28.38} & \multicolumn{1}{c}{46.74} & \multicolumn{1}{c}{19.92}
                  & \multicolumn{1}{c}{31.3} \\
WSLAT-Ens  \cite{rochan2015weakly}      & \multicolumn{1}{c}{78.57} & \multicolumn{1}{c}{63.37} & \multicolumn{1}{c}{66.36} & \multicolumn{1}{c}{56.35} & \multicolumn{1}{c}{19.67}
                  & \multicolumn{1}{c}{\textbf{82.26}} & \multicolumn{1}{c}{74.75} & \multicolumn{1}{c}{\textbf{69.13}} & \multicolumn{1}{c}{22.47} & \multicolumn{1}{c}{72.34}
                  & \multicolumn{1}{c}{31.00} & \multicolumn{1}{c}{62.95} & \multicolumn{1}{c}{74.91} & \multicolumn{1}{c}{78.37} & \multicolumn{1}{c}{\textbf{48.61}}
                  & \multicolumn{1}{c}{29.39} & \multicolumn{1}{c}{64.58} & \multicolumn{1}{c}{36.24} & \multicolumn{1}{c}{\textbf{75.86}} & \multicolumn{1}{c}{\textbf{69.53}}
                  & \multicolumn{1}{c}{58.84} \\ \hline
B-MSD-VGG16  & \multicolumn{1}{c}{60.93} & \multicolumn{1}{c}{68.24} & \multicolumn{1}{c}{52.25} & \multicolumn{1}{c}{34.04}  & \multicolumn{1}{c}{38.55}
                  & \multicolumn{1}{c}{69.04} & \multicolumn{1}{c}{75.03} & \multicolumn{1}{c}{58.43} & \multicolumn{1}{c}{12.41}  & \multicolumn{1}{c}{74.66}
                  & \multicolumn{1}{c}{36.12} & \multicolumn{1}{c}{58.37} & \multicolumn{1}{c}{55.78} & \multicolumn{1}{c}{83.13} & \multicolumn{1}{c}{40.81}
                  & \multicolumn{1}{c}{37.00} & \multicolumn{1}{c}{73.20} & \multicolumn{1}{c}{41.40} & \multicolumn{1}{c}{47.91} & \multicolumn{1}{c}{42.29}
                  & \multicolumn{1}{c}{52.97} \\
OOM-MSD-AlexNet   & \multicolumn{1}{c}{86.25} & \multicolumn{1}{c}{66.67} & \multicolumn{1}{c}{66.37} & \multicolumn{1}{c}{57.45} & \multicolumn{1}{c}{29.01}
                  & \multicolumn{1}{c}{66.50} & \multicolumn{1}{c}{71.48} & \multicolumn{1}{c}{66.86} & \multicolumn{1}{c}{20.80} & \multicolumn{1}{c}{73.97}
                  & \multicolumn{1}{c}{30.04} & \multicolumn{1}{c}{67.44} & \multicolumn{1}{c}{72.11} & \multicolumn{1}{c}{87.15} & \multicolumn{1}{c}{29.40}
                  & \multicolumn{1}{c}{31.87} & \multicolumn{1}{c}{78.35} & \multicolumn{1}{c}{47.85} & \multicolumn{1}{c}{65.40} & \multicolumn{1}{c}{56.63}
                  & \multicolumn{1}{c}{58.58}  \\
Ours-MSD-AlexNet  & \multicolumn{1}{c}{85.00} & \multicolumn{1}{c}{72.16} & \multicolumn{1}{c}{66.37} & \multicolumn{1}{c}{66.49} & \multicolumn{1}{c}{38.17}
                  & \multicolumn{1}{c}{73.60} & \multicolumn{1}{c}{78.71} & \multicolumn{1}{c}{66.28} & \multicolumn{1}{c}{22.90} & \multicolumn{1}{c}{79.45}
                  & \multicolumn{1}{c}{27.38} & \multicolumn{1}{c}{72.09} & \multicolumn{1}{c}{78.91} & \multicolumn{1}{c}{87.55} & \multicolumn{1}{c}{15.75}
                  & \multicolumn{1}{c}{39.56} & \multicolumn{1}{c}{81.44} & \multicolumn{1}{c}{41.94} & \multicolumn{1}{c}{63.50} & \multicolumn{1}{c}{62.72}
                  & \multicolumn{1}{c}{61.00}  \\
Ours-MSD-VGG16    & \multicolumn{1}{c}{85.83} & \multicolumn{1}{c}{\textbf{77.25}} & \multicolumn{1}{c}{73.57} & \multicolumn{1}{c}{49.47} & \multicolumn{1}{c}{\textbf{66.41}}
                  & \multicolumn{1}{c}{76.14} & \multicolumn{1}{c}{84.36} & \multicolumn{1}{c}{74.13} & \multicolumn{1}{c}{32.87} & \multicolumn{1}{c}{\textbf{86.99}}
                  & \multicolumn{1}{c}{19.01} & \multicolumn{1}{c}{\textbf{84.19}} & \multicolumn{1}{c}{\textbf{85.03}} & \multicolumn{1}{c}{90.36} & \multicolumn{1}{c}{13.37}
                  & \multicolumn{1}{c}{34.43} & \multicolumn{1}{c}{\textbf{88.66}} & \multicolumn{1}{c}{44.09} & \multicolumn{1}{c}{71.86} & \multicolumn{1}{c}{67.38}
                  & \multicolumn{1}{c}{65.27}  \\
Ours-MSD-Ens      & \multicolumn{1}{c}{\textbf{89.17}} & \multicolumn{1}{c}{75.69} & \multicolumn{1}{c}{\textbf{75.08}} & \multicolumn{1}{c}{\textbf{66.49}} & \multicolumn{1}{c}{58.78}
                  & \multicolumn{1}{c}{78.17} & \multicolumn{1}{c}{\textbf{88.89}} & \multicolumn{1}{c}{66.86} & \multicolumn{1}{c}{28.15} & \multicolumn{1}{c}{86.30}
                  & \multicolumn{1}{c}{29.66} & \multicolumn{1}{c}{83.49} & \multicolumn{1}{c}{83.33} & \multicolumn{1}{c}{\textbf{92.77}} & \multicolumn{1}{c}{23.68}
                  & \multicolumn{1}{c}{40.29} & \multicolumn{1}{c}{85.57} & \multicolumn{1}{c}{\textbf{48.92}} & \multicolumn{1}{c}{70.34} & \multicolumn{1}{c}{68.10}
                  & \multicolumn{1}{c}{\textbf{66.79}}  \\ \hline
\end{tabular}
}
\vspace{-1em}
\end{center}
\end{table*}

\subsubsection{Implementation Details}
\label{section:cross-dataset-imple}

In cross-dataset detection task, the most of the experimental settings (learning rates, nms threshold, \etc) are as same as the ones in intra-dataset detection case (Section \ref{section:intra-dataset-imple}). The only difference is that, when the detection models in Ours-MSD and the three baselines are learned, we use five image scales $\left \{480,576,688,864,1200 \right \}$ for both training and testing as an additional form of data augmentation. This multi-scale training/testing strategy is widely-used in recent fully/weakly supervised detection methods \cite{bilen2016weakly,dai2016r,li2016weakly} and has proven effective on PASCAL VOC dataset.

\subsubsection{Evaluation on PASCAL VOC 2007}
Our results for each class on PASCAL VOC 2007 are reported in Table \ref{tab:voc-cross-mAP} (mAP) and Tabel \ref{tab:voc-cross-corloc} (CorLoc). The first compartment in both tables shows the results obtained by state-of-the-art WSD methods \cite{bilen2016weakly,li2016weakly} and B-WSD baseline, which are trained on the weak categories (20 categories in PASCAL VOC) only. The second compartment reports the results of the MSD methods that leverage extra strong categories (180 categories in ILSVRC2013) for training. Additionally, in Table \ref{tab:voc-cross-mAP}, the performance of fully supervised Fast RCNN detectors is listed in the third compartment. The Fast RCNN detectors are trained without bounding box regression and the results are cited from the \textit{experiment logs} released by Fast RCNN. \footnote{\url{https://dl.dropboxusercontent.com/s/q4i9v66xq9vhskl/fast_rcnn_experiments.tgz?dl=0}}.

As shown in Table \ref{tab:voc-cross-mAP} and Table \ref{tab:voc-cross-corloc}, using a single AlexNet model, Ours-MSD achieves huge improvements in mAP (41.17 versus 23.87 percent) and in CorLoc (61.00 versus 41.35 percent) compared with the B-WSD baseline. This performance also significantly exceeds the state-of-the-art WSD results \cite{diba2016weakly,kantorov2016contextlocnet} (41.77 versus 37.3/36.3 percent). When compared with OOM-MSD, our method with robust objectness knowledge also shows superiority (41.77 versus 37.65 percent). When Ours-MSD is trained with the deeper VGG16 detectors, the result is improved to 47.50 percent, which also largely outperforms the B-WSD baseline (25.03 percent), the B-MSD baseline (33.53 percent) and previous WSD results \cite{diba2016weakly,li2016weakly} (42.8/39.3 percent) that also adopt VGG16 detectors. Similar to WSDDN \cite{bilen2016weakly}, our results can be further improved by combing multiple models. The ensemble model used in our method (Ours-MSD-Ens) is obtained by simply summing up the scores of AlexNet detector (Ours-MSD-AlexNet) and VGG16 detector (Ours-MSD-VGG16) and it finally achieves 49.82 percent, which outperforms the ensemble results in WSDDN by a large margin (49.82 versus 39.3 percent). Also, similar to \cite{li2016weakly}, we use the obtained ensemble model to select top-scoring regions (select one highest-score region from each image of each category) as pseudo ground truth boxes to train a supervised VGG16 Fast RCNN detector \cite{girshick2015fast} with no bounding box regression. Further improvements can be obtained with this process (51.08 versus 49.82 percent). Finally, as shown in Table \ref{tab:voc-cross-mAP}, even when compared with fully supervised Fast RCNN detectors, our MSD methods can achieve comparable detection results.

Then we compare the proposed method with the state-of-the-art MSD method, WSLAT \cite{rochan2015weakly}, which also leverages the strong categories of ILSVRC2013 to support the WSD learning on PASCAL VOC 2007. The proposed method is compared with WSLAT in terms of CorLoc (\%) in Table \ref{tab:voc-cross-corloc}, since WSLAT does not report their mAP results on VOC 2007 \textit{test} set.  It is noted that WSLAT has \textit{three} variants in \cite{rochan2015weakly}: 1) a WSD model directly trained on weak categories (denoted as WSLAT$_{\mathit{weak}}$); 2) a transfer model (denoted as WSLAT$_{\mathit{trans}}$) built on both strong and weak categories leveraging their semantic relationships, and 3) ensemble model (denoted as WSLAT-Ens) by combining the above two. Our method outperforms \textit{all} the three variants. In particular, WSLAT$_{\mathit{trans}}$ is mostly close to us. However, its reported CorLoc value is 31.3 percent \cite{rochan2015weakly}, far below that of our method (61.00 percent). Only by combining a high-performance WSD model (WSLAT$_{\mathit{weak}}$, 53.00 percent), does the WSLAT obtain an acceptable result (WSLAT-Ens, 58.84 percent).

\begin{table}[t]
\renewcommand{\arraystretch}{1.2}
\captionsetup{justification=centering}
\caption{\label{tab:voc-cross-1012}\\Object Detection Performance (mAP \%) on PASCAL VOC 2010 \textit{test} Set and VOC 2012 \textit{test} Set.}
\vspace{-1em}
\begin{center}
\small
\begin{tabular}{p{4.0cm}  p{1.5cm}<{\centering} p{1.5cm}<{\centering}}
\hline
         Method                               & VOC 2010 & VOC 2012       \\ \hline
         \hline
weakly supervised                             &          &                \\
B-WSD-AlexNet                                 &   25.25       &    24.65            \\
B-WSD-VGG16                                   &    21.51     &    21.44            \\ \hline
(OM+MIL)+FRCN-AlexNet \cite{li2016weakly}     & 21.4 & 22.4 \\
(OM+MIL)+FRCN-VGG16 \cite{li2016weakly}       & 30.7 & 29.1 \\ \hline
ContextLocNet \cite{kantorov2016contextlocnet}& -    & 35.3 \\ \hline
WCCN-AlexNet \cite{diba2016weakly}            & 28.8 & 28.4 \\
WCCN-VGG16 \cite{diba2016weakly}            & 39.5 & 37.9 \\ \hline
\hline
mixed supervised                             &          &  \\
B-MSD-VGG16                                  & 33.92    & 33.93 \\
OOM-MSD-AlexNet                              & 35.42    & 35.43 \\
Ours-MSD-AlexNet                             & 37.98    & 38.12 \\
Ours-MSD-VGG16                               & \textbf{42.87}    & \textbf{43.42} \\ \hline
\end{tabular}
\end{center}
\vspace{-1em}
\end{table}

\subsubsection{Evaluation on PASCAL VOC 2010 and VOC 2012}
In this section, our MSD method (Ours-MSD) is compared with the baseline methods and other state-of-the-art WSD methods on VOC 2010 and VOC 2012. The detection results (mAP \%) are reported in Table \ref{tab:voc-cross-1012}. In general, the detection performance of VOC 2010 and VOC 2012 is lower than that of VOC 2007. When compared with WSD methods, Ous-MSD significantly outperforms B-WSD baseline and the state-of-the-art results \cite{diba2016weakly,kantorov2016contextlocnet,li2016weakly} with both AlexNet detectors and VGG16 detectors. When the two objectness transfer methods are compared, Ous-MSD also surpasses OOM-MSD. Finally, Ours-MSD outperforms B-MSD baseline by a large margin on both VOC 2010 and VOC 2012 datasets. It conforms the superiority of the robust objectness transfer approach over the straightforward fine-tuning approach.

\begin{table}[t]
\renewcommand{\arraystretch}{1.2}
\captionsetup{justification=centering}
\caption{\label{tab:voc-cross-onlyobjects}\\Object Detection Performance (mAP) on PASCAL VOC 2007.}
\vspace{-1em}
\begin{center}
\normalsize
\begin{tabular}{p{3cm} p{2.0cm}<{\centering}}
\hline
         Method                               & mAP       \\ \hline
B-WSD                                 &   23.87                \\ \hline
MSD-no-distractor                     & 25.95                   \\
Ours-MSD                             & 41.77 \\ \hline
\end{tabular}
\end{center}
\footnotesize
\begin{tablenotes}
\item [1] \textit{``MSD-no-distractor'' indicates the alternative baseline that only learns 20 object categories from the selected object regions (top 15 percent).}
    \vspace{-1em}
\end{tablenotes}
\end{table}

\subsection{Ablation Studies}
In this section, we conduct some ablation experiments to illustrate the effectiveness of our robust objectness transfer MSD approach. Without loss of generality, the comparisons are performed on cross-dataset detection task and trained with the AlexNet model. All the experiments follow the same settings mentioned in Section \ref{section:cross-dataset-imple} (learning rates, nms threshold, multi-scale strategies, \etc).

\subsubsection{Is Learning the Concept of Distractors Necessary for WSD?}
To explore the necessity of modelling distractors in WSD, we propose an alternative baseline, MSD-no-distractor and compare it with the proposed method (Ours-MSD). MSD-no-distractor also utilizes the objectness knowledge but aims to learn object categories only. Specially, MSD-no-distractor first uses the obtained objectness model to score the regions in each weakly labeled image. Then it selects the top 15 percent of regions as the ``object regions'', which is similar to Ours-MSD. The difference is that MSD-no-distractor only utilizes the selected top 15 percent ``object regions'' to train a 20-class (20 categories of PASCAL VOC) WSD model; while Ours-MSD utilizes both the top 15 percent ``object regions'' and the last 85 percent ``non-object regions'' to train a 20+1-class objectness-aware detection model.

The performance comparison of the three methods (B-WSD, Ours-MSD and MSD-no-distractor) is shown in Table \ref{tab:voc-cross-onlyobjects}. We can see that the performance of MSD-no-distractor is only slightly better than that of B-WSD (25.95 versus 23.87 percent). Compared with B-WSD, MSD-no-distractor has already largely reduced the search space for object categories (from 100 percent regions used in B-WSD to the selected 15 percent regions used in MSD-no-distractor), but it still cannot distinguish the objects from distractors. When the trained detectors saw a distractor, \eg, a \textit{cat face}, the MSD-no-distractor cannot recognize it as a false detection due to the missing of distractor concept and the obtained improvements are quite small. Only when we learn the distractor concept together with the object categories (Ours-MSD), does the detector correctly distinguish between objects and distractors and achieve remarkable improvements (41.77 versus 23.87 percent).

\begin{figure}[]
\begin{center}
      \includegraphics[width=0.9\linewidth]{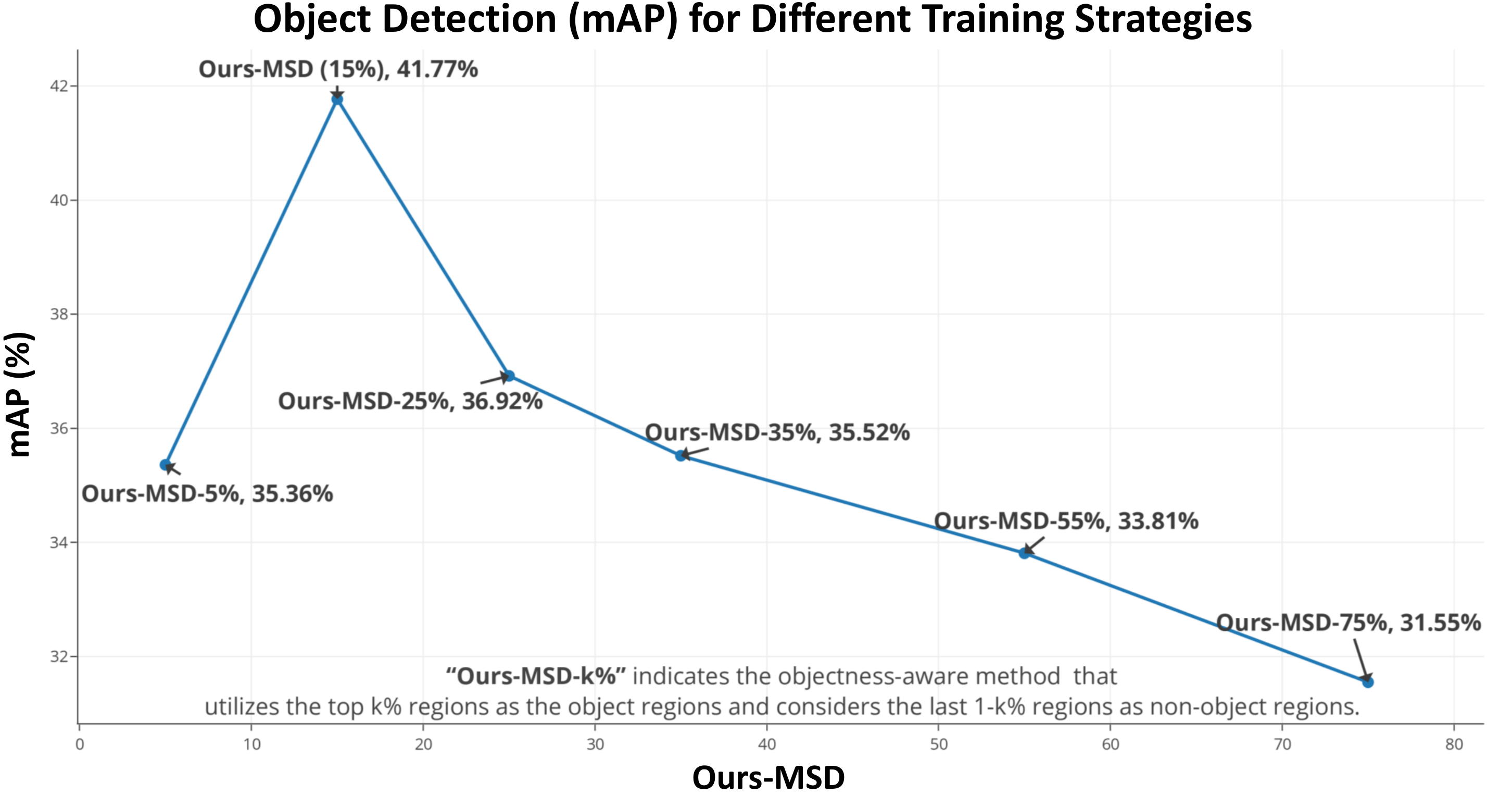}
\end{center}
   \caption{\label{fig:voc-cross-topk}Object Detection performance (mAP) on PASCAL VOC 2007. ``Ours-MSD-k\%'' indicates the objectness-aware method that utilizes the top k\% regions as the object regions and considers the last 1-k\% regions as non-object regions.}
   \vspace{-1em}
\end{figure}

\subsubsection{The Effect of the Quantity of Selected Object Regions}
In Ours-MSD, we select the top 15 percent of regions as object regions to train the objectness-aware detection models. In this ablation experiment, we analyse the influence of selected object regions's quantity. We apply the same domain-invariant objectness model to re-rank the regions in weakly labeled images. Then we use top 5, top 25, top 35, top 55, top 75 percent of regions as object regions respectively meanwhile utilize the remaining regions as non-object regions to train the objectness-aware detection models. The results are shown in Fig. \ref{fig:voc-cross-topk}. It can be seen that when the number of selected object regions are small (\eg, top 5 percent), the selected object bags do not contain enough positive regions (IoU$\geq$0.5) in images (The detailed recall numbers of positive regions can be found in Fig. \ref{fig:recall}). Thus the performance improvement of subsequent detection models is limited (Ours-MSD-5, 35.36 percent). With more regions selected as objects (from top 5 to top 15 percent), the recall of positive regions in selected object bags gradually increases, and the performance of detection models also improves (from 35.36 to 41.77 percent). But when more regions are chosen as objects, the improvements of the recall are relatively small. Moreover, the large number of selected regions would bring in lots of false positives and decrease the performance of detectors.

\subsubsection{Discrepant Domains Transfer (Natural Objects versus Man-made Objects)}
In this section, we aim to apply our objectness transfer approach between more discrepant domains. We select 8 \textit{natural object} categories from PASCAL VOC 2007 and construct set $\mathcal{W}$ with images of natural objects (1,929 images). Then all the \textit{man-made} objects are selected from ILSVRC2013 detection (139 categories in total) and their images construct set $\mathcal{S}$. The proposed method is compared with the B-WSD method and the results are shown in Table \ref{tab:voc-manvsna}. It can be seen that the performance of Ours-MSD significantly outperforms the B-WSD (31.08 versus 21.17 percent). Even when the strong categories and weak categories come from more discrepant domains, \ie, \textit{man-made} versus \textit{natural}, our objectness transfer approach is still effective to improve the WSD performance.

\begin{table}[t]
\renewcommand{\arraystretch}{1.15}
\captionsetup{justification=centering}
\caption{\label{tab:voc-manvsna}\\Object Detection Performance (mAP) on PASCAL VOC 2007 \textit{test} Set.}
\vspace{-1em}
\begin{center}
{
\begin{tabular}{|p{1.4cm}|p{0.3cm}<{\centering} p{0.25cm}<{\centering} p{0.25cm}<{\centering}
                          p{0.3cm}<{\centering} p{0.4cm}<{\centering}p{0.5cm}<{\centering}
                          p{0.4cm}<{\centering}p{0.6cm}<{\centering}
                          |p{0.5cm}<{\centering}|}
\hline
Method  & bird & cat  & cow  & dog & horse & person & plant & sheep & mAP \\ \hline
B-WSD  & 21.5 & 20.8 & 25.9 & 26.7 & 23.5 & 18.3 & 12.8 & 19.9 & 21.17 \\ \hline
Ours-MSD  & 40.7 & 37.0 & 37.6 & 23.1 & 45.1 & 18.6 & 17.9 & 28.6 & 31.08 \\ \hline
\end{tabular}
}
\end{center}
\footnotesize
\begin{tablenotes}
\item [1] \textit{The mAP is computed over 8 \textit{natural object} categories.}
    \vspace{-1em}
\end{tablenotes}
\end{table}

\subsubsection{Comparisons with Other Objectness Detectors}
\begin{figure}[]
\begin{center}
      \includegraphics[width=0.9\linewidth]{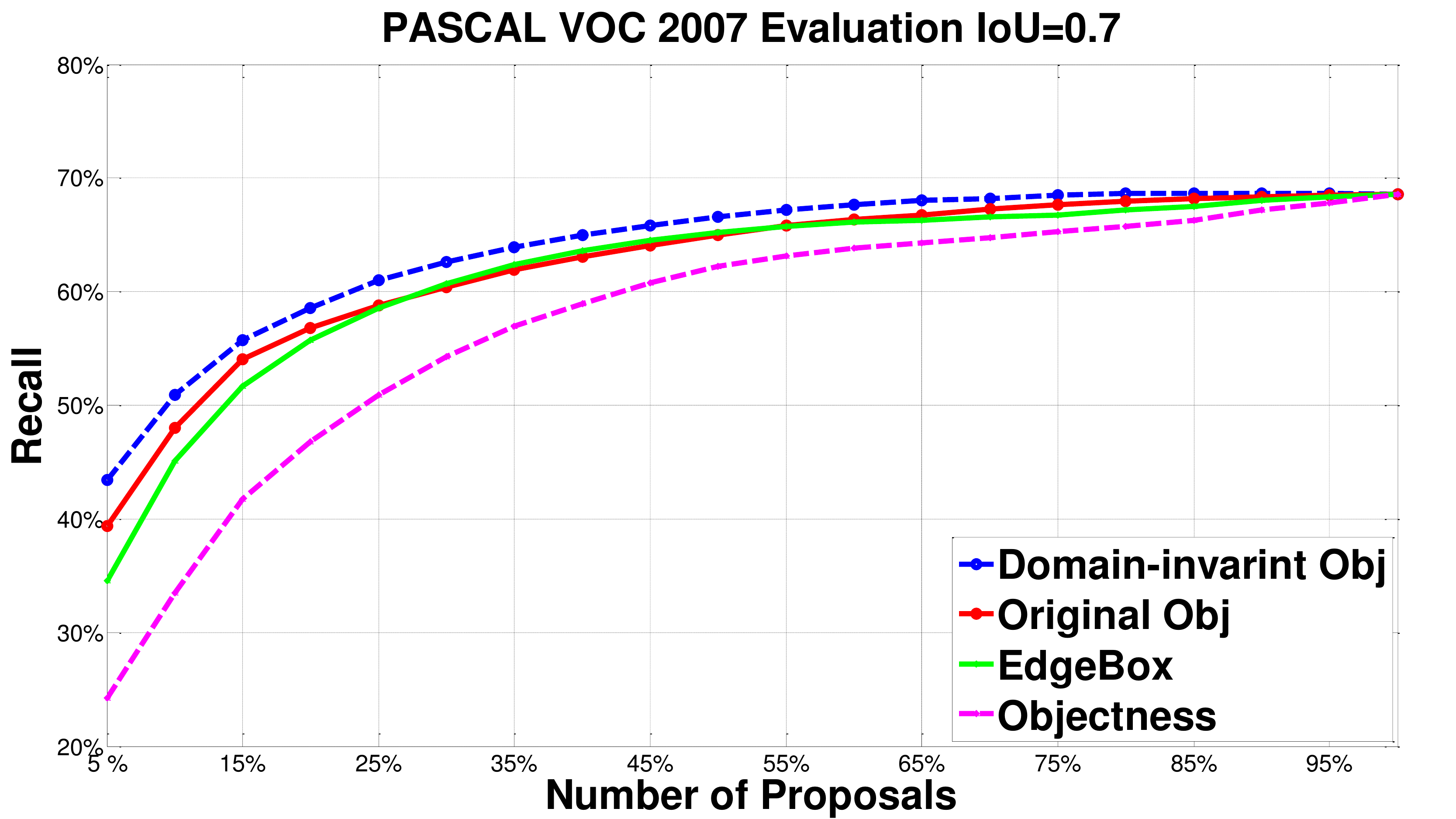}
\end{center}
   \caption{The recall rates on PASCAL VOC 2007 \textit{trainval} set. We compute the recall using the code provided by \cite{zitnick2014edge}. The four curves join at 100\% region proposals because we re-rank the existing selective search windows.}
\label{fig:recall}
\end{figure}

Further experiments are conducted to compare our domain-invariant objectness model with other objectness/proposal methods for \textit{objectness} and \textit{object instance} detection on PASCAL VOC 2007. Four models are considered: Objectness \cite{alexe2012measuring}, EdgeBox \cite{zitnick2014edge} \footnote{The official code of EdgeBox \cite{zitnick2014edge} uses the edge-driven \textit{objectness} measure to score sliding windows and finally outputs the selected windows with their \textit{objectness} scores. To calculate the \textit{objectness} score for a specific given box, we use its modified code provided by \cite{cinbis2017weakly}.}, original objectness model (OOM-MSD, Section \ref{section:baseline}) and our domain-invariant objectness model (Ours-MSD, Section \ref{section:objectness}). For each model, we re-rank the selective search windows based on their \textit{objectness} scores, and compute the \textit{recall} for different percentage of the proposals (\ie, percentage of windows considered containing an object instance) when IoU=0.7. The results are shown in Fig. \ref{fig:recall}. It can be seen that our domain-invariant objectness model outperforms existing objectness models (\ie, Objectness \& EdgeBox) and the original objectness model in all cases. This confirms that our domain-invariant objectness model is a better objectness \textit{detector}, which accounts for the better performance of the proposed domain-invariant objectness.

To test the effect of different objectness models on \textit{object instance} detection, for each of them, we select the top 15 percent of the re-ranked selective search windows as ``objects'' to train the objectness-aware detection model. The detection results are shown in Table \ref{tab:voc-objectness}. We can see that when we use existing objectness models, the subsequent detection models (Objectness-MSD \& EdgeBox-MSD) obtain significantly lower performance than the ones using the CNN-based objectness models, OOM-MSD \& Ours-MSD (10.99\%\&19.78\% \vs 37.65\%\&41.77\%). When the two objectness models are compared, Ours-MSD outperforms OOM-MSD.

It is quite a surprising result that the performance of EdgeBox-MSD is much lower than that of OOM-MSD \& Ours-MSD. As shown in Fig. \ref{fig:recall}, the recall numbers at 15 percent for three methods (EdgeBox \& original objectness \& domain-invariant objectness) are very close, while the detection performance of subsequent detection models differs greatly (19.78 \& 37.65 \& 41.77 percent). So what makes such a contradiction?

\begin{table}[]
\renewcommand{\arraystretch}{1.2}
\captionsetup{justification=centering}
\caption{\label{tab:voc-objectness}\\Object Detection Performance (mAP) on PASCAL VOC 2007.}
\vspace{-1em}
\begin{center}
\normalsize
\begin{tabular}{p{4.0cm} p{2cm}<{\centering}}
\hline
         Method                               & mAP       \\ \hline
Objectness-MSD \cite{alexe2012measuring}                                 &   10.99               \\
EdgeBox-MSD \cite{zitnick2014edge}                    & 19.78                  \\ \hline \hline
OOM-MSD                             & 37.65 \\
Ours-MSD                            & 41.77 \\ \hline
\end{tabular}
\end{center}
\footnotesize
\begin{tablenotes}
\item [1] \textit{``Objectness-MSD'' \& ``EdgeBox-MSD'' indicate the methods that utilize objectness detector (Objectness \& EdgeBox) to score selective search windows and then train objectness-aware detection models.}
    \vspace{-1em}
\end{tablenotes}
\end{table}

\begin{figure}[]
\begin{center}
      \includegraphics[width=0.9\linewidth]{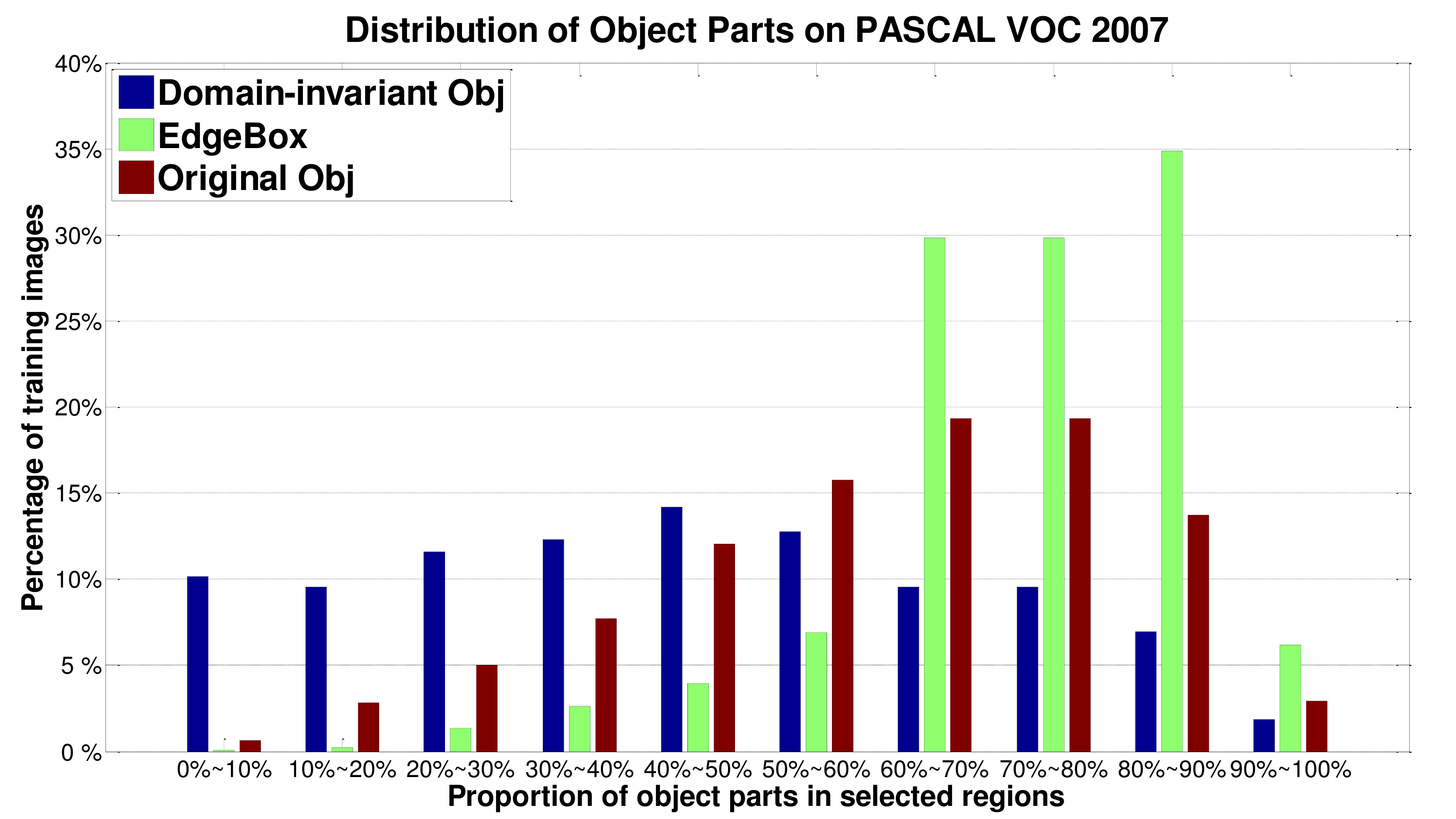}
\end{center}
   \caption{The distribution of \textit{object parts} for three methods evaluated on PASCAL VOC 2007 \textit{trainval} set. The \textit{object parts} are the regions whose IoUs with ground truths are in the interval $\left( 0.0,0.5 \right)$. The \textit{object parts} are difficult to reject in weakly supervised setting.}
\label{fig:distribution}
\end{figure}

We explore the reasons from the pitfall in weakly supervised detection (WSD) as we mentioned in Section \ref{sec:intro}. The regions in an image can be divided into three types according to their IoUs with ground truths: \textit{positive objects} (IoU$\geq$0.5), \textit{object parts} (0$<$IoU$<$0.5) and \textit{backgrounds} (IoU=0). For WSD, the \textit{positive objects} and \textit{backgrounds} are easily to be distinguished in most cases and the main difficulty is how to separate \textit{positive objects} from \textit{object parts}. In selected object regions (\ie, the top 15 percent of regions in our experimental setting), few \textit{positive objects} or excess \textit{object parts} would both hurt the WSD performance. The recall rates in Fig. \ref{fig:recall} only show the number of \textit{positive objects} in selected regions and do not consider the \textit{object parts}. Thus it cannot roundly reflect the effectiveness of these objectness detectors in WSD. To address this issue, we conduct another experiment to visualize the distribution of \textit{object parts} for three methods. The results are shown in Fig. \ref{fig:distribution}.

\begin{figure*}[]
\begin{center}
   \includegraphics[width=0.9\linewidth]{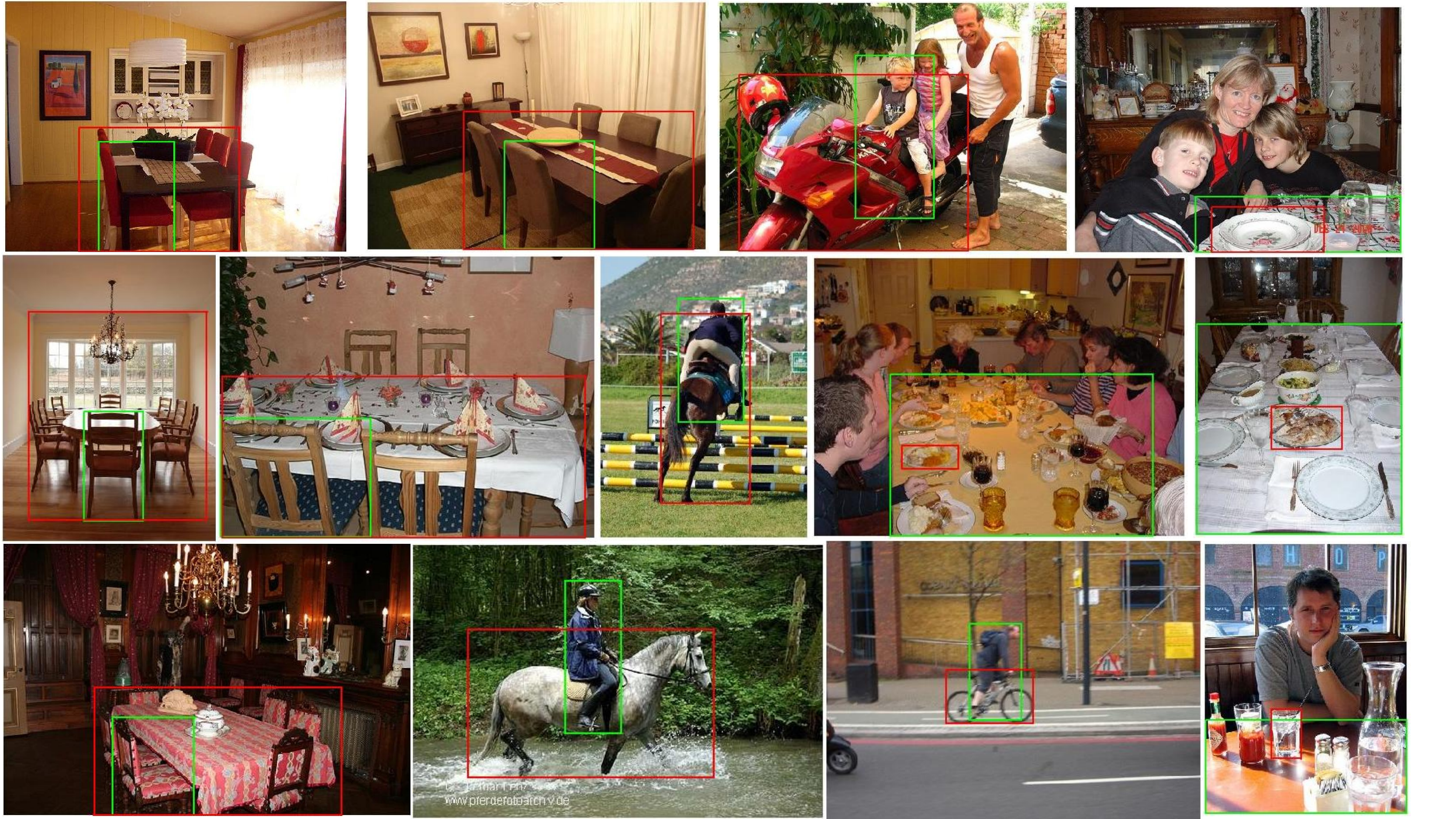}
\end{center}
   \caption{Sample failure detection results on PASCAL VOC 2007 \textit{test} set for class \textit{chair}, \textit{table} and \textit{person}. Green rectangles are ground truth boxes, and red rectangles indicate failure detections (IoU$<$0.5).}
\label{fig:visual}
\end{figure*}

In Fig. \ref{fig:distribution}, the x-coordinate stands for the proportion of \textit{object parts} in selected regions and the y-coordinate stands for the percentage of images having the corresponding proportion range of object parts in all training images. For example, the x-coordinate of the first blue bar is (0$\sim$10 percent), indicating that ``for a weakly labeled image with $n$ ($n$=2000 for instance) region proposals, we select the top 15\% regions as objects ($2000 \times 0.15 = 300$ regions); in the selected 300 regions, the 0\%$\sim$10\% of these regions (0$\sim$30 regions) are actually \textit{object parts}''. Meanwhile, the y-coordinate of the first blue bar, 10.18 percent, means that ``in all 5011 training images, there are 10.18 percent of the images (around 500 images), in each of which 0$\sim$10 percent of the selected regions are \textit{object parts}.'' As shown in Fig. \ref{fig:distribution}, the mode of distribution curve of \textit{object parts} for EdgeBox (green bars) is obviously on the \textbf{right} to the ones of our objectness models (red \& blue bars), which means there are more \textit{object parts} in selected regions for EdgeBox. It is noted that EdgeBox only utilizes low-level cues (\ie, contour information) to measure objectness, which has limited capability of rejecting \textit{object parts} in images. Considering a typical case that many \textit{object parts}, such as \textit{cat face}, also have closed contours and can be recognized as \textit{objects} with EdgeBox. In contrast, our objectness model \textit{learns} to capture the concept of ``complete objectness'' from lots of annotated data, which results in less \textit{object parts}.

We can also observe that the \textit{object parts} in domain-invariant objectness models are also less than the original objectness models. The reason is that the domain-invariant models would include more \textit{backgrounds} in selected regions than the original objectness models. One overlooked fact in PASCAL VOC is that the images in VOC contain lots of ``non-target objects'' that do not belong to VOC 20 categories \cite{chavali2016object}. That is, the \textit{backgrounds} regions also contain a lot of ``complete objects'' belonging to non-target categories. Moreover, when we train domain-invariant objectness models, all regions in images of target domain (PASCAL VOC) are randomly sampled to learn domain-invariant features. Thus, the learned domain-invariant objectness would be robust to not only VOC 20 categories but also the non-target categories. Finally, the ``non-target objects'' in \textit{backgrounds} are more likely to be recognized as \textit{objects} in domain-invariant objectness models, which results in higher objectness scores for \textit{backgrounds} and leads to fewer \textit{object parts} in selected top 15 percent regions. Considering both \textit{positive objects} (Fig. \ref{fig:recall}) and \textit{object parts} (Fig. \ref{fig:distribution}), our experiments clearly confirm the superiority of our domain-invariant objectness model for both \textit{objectness} and \textit{object instance} detection, especially in weakly supervised settings.

\subsection{Error Analysis}
Though our method achieves outstanding performance for many categories, its performance is still poor for classes such as ``chair'', ``table'' and ``person''. For analysis, we show failure detection results on VOC 2007 \textit{test} in Fig. \ref{fig:visual}. We can see that, for ``chair'' images, multiple \textit{chairs} often get close together and the \textit{chairs} typically co-appear with a \textit{table}. In this case, it is difficult to figure out single complete \textit{chair} from images. The detectors would prefer to select the whole \textit{table}, as the \textit{table} is the most likely ``object''. The similar situation also exists in ``table'' images. The \textit{tables} often appear together with other categories, such as \textit{bottle}, \textit{plate} and \textit{person}. The exact closed contours for such tables cannot be clearly and easily confirmed, and the complete \textit{bottles}, \textit{plates}, or even \textit{pizzas} would be more easily recognized as objects in such table images. The main kind of failure detections of \textit{person} images are caused by the multiple category setting. In PASCAL VOC, some images contain more than one categories, and for example in many person images, the person appears together with \textit{bicycle}, \textit{horse}, or \textit{motorbike}. In such a situation, the selected object regions are considered as positive for both \textit{person} and \textit{horse} categories, and the detectors cannot distinguish between the \textit{person} object and the \textit{horse} object. This issue caused by multiple category setting is, to some extent, intrinsic in weakly supervised settings where only image category labels are available.

\section{Conclusion}

In this paper, we consider mixed supervised detection (MSD), which aims to leverage the existing fully labeled categories to localize objects of new categories with weak labels only. The weakly supervised detection of new objects does not require expensive bounding box annotations, and satisfactory detection solutions can be obtained by exploiting the existing fully labeled categories. These characteristics make MSD be a practically important problem.

In MSD, the existing fully labeled categories have no overlap with new categories. Thus, the key issue to be solved is how to learn the transferable and robust knowledge from the existing categories to assist the detection on new categories. Previous MSD works \cite{hoffman2014lsda,hoffman2015detector,rochan2015weakly,tang2016large} transfer the learned object detectors from the existing categories to new categories following some hand-crafted strategies. In contrast, the proposed robust objectness transfer approach automatically \textit{learns} the domain-invariant knowledge, and the proposed objectness-aware detection model further utilizes the learned objectness knowledge to distinguish the objects from distractors. The state-of-the-art object detection performance has been achieved on benchmarking datasets, which confirms the superiority of our proposed method.

\section{Acknowledgements}
This work is funded by the National Key Research and Development Program of China (Grant 2016YFB1001004 and Grant 2016YFB1001005), the National Natural Science Foundation of China (Grant 61673375, Grant 61721004 and Grant 61403383) and the Projects of Chinese Academy of Sciences (Grant QYZDB-SSW-JSC006 and Grant 173211KYSB20160008).

\bibliographystyle{IEEEtran}
\bibliography{mybibfile}

\begin{IEEEbiography}[{\includegraphics[width=1.25in,height=1.25in,clip,keepaspectratio]{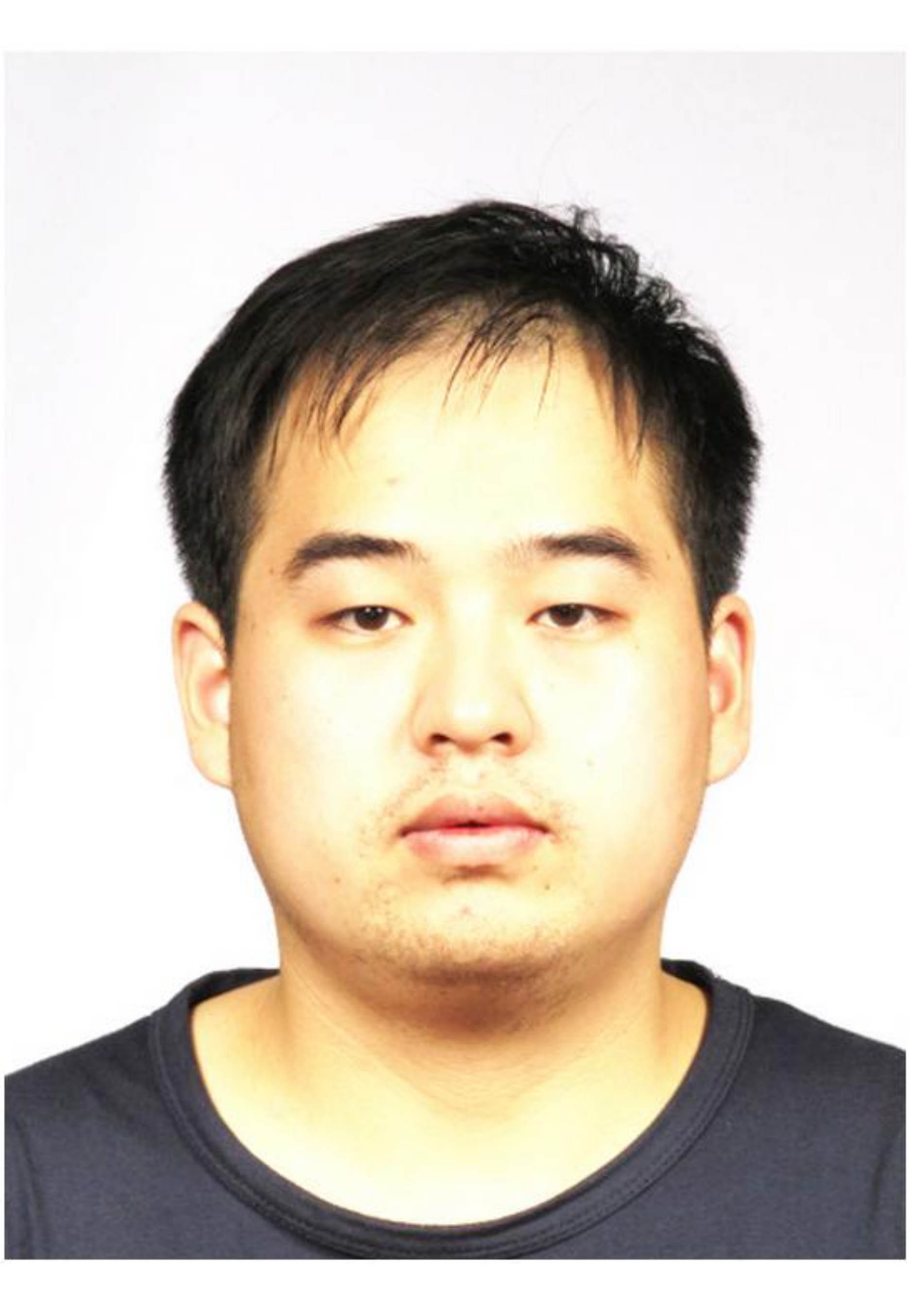}}]{Yan Li}
    received the B.Eng degree in Automation from Tsinghua University, Beijing, China, in 2014. He is currently working toward the PhD degree in the Center for Research on Intelligent Perception and Computing (CRIPAC), Institute of Automation, Chinese Academy of Sciences (CASIA), Beijing, China. His research interests include computer vision, object detection, weakly supervised learning and few shot learning. He is a student member of the IEEE.
\end{IEEEbiography}

\begin{IEEEbiography}[{\includegraphics[width=1.25in,height=1.25in,clip,keepaspectratio]{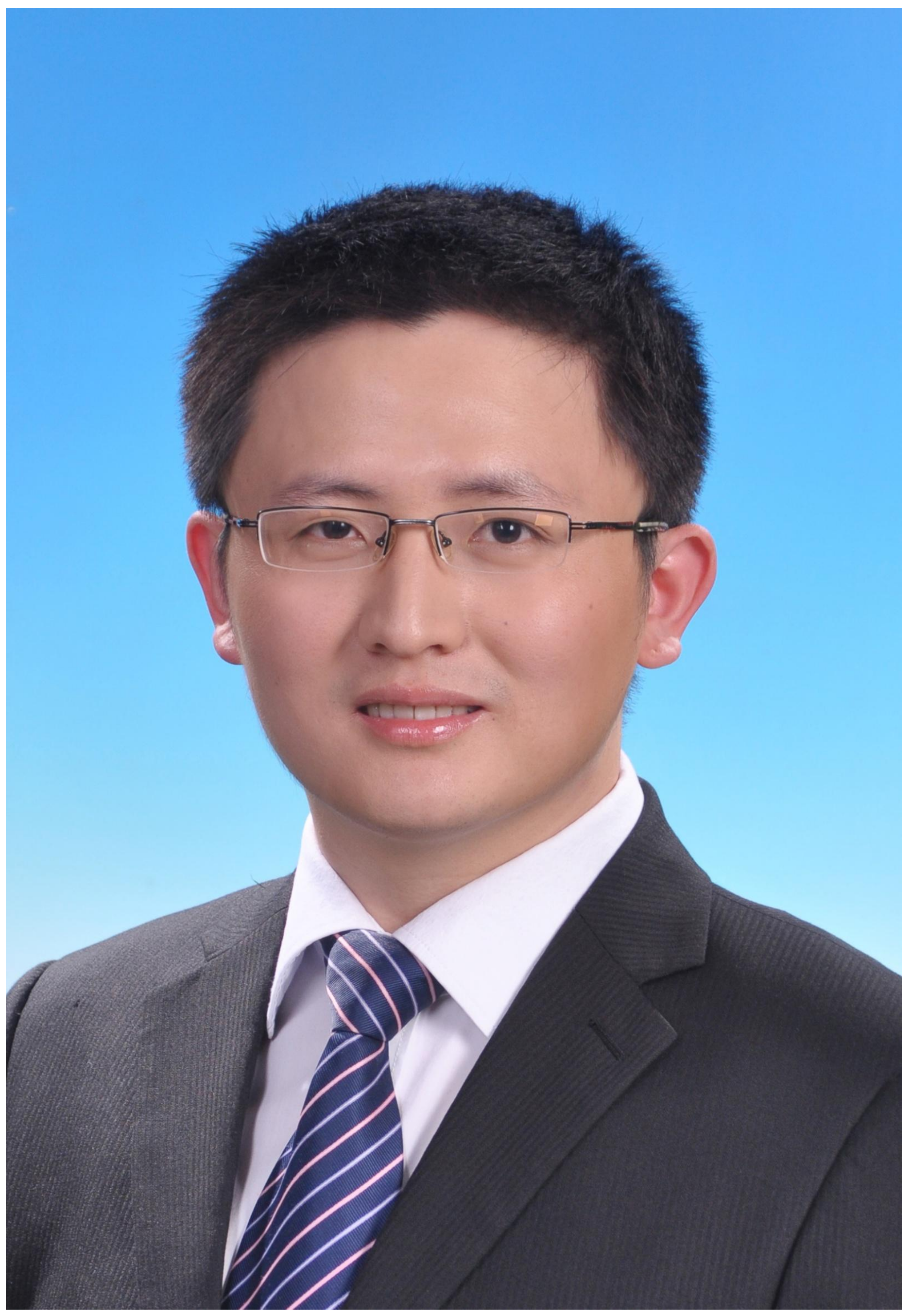}}]{Junge Zhang} received his B.S. degree from China University of Geosciences (CUG) in 2008. He received his PhD degree in Pattern Recognition and Intelligent Systems from the Institute of Automation, Chinese Academy of Sciences (CASIA) in 2013. In July 2013, Dr. Zhang joined the Center for Research on Intelligent Perception and Computing (CRIPAC). Now he serves as a Special-appointed Professor. His major research interests include computer vision, pattern recognition, deep learning and general artificial intelligence. He is a member of the IEEE.
\end{IEEEbiography}

\begin{IEEEbiography}[{\includegraphics[width=1.25in,height=1.25in,clip,keepaspectratio]{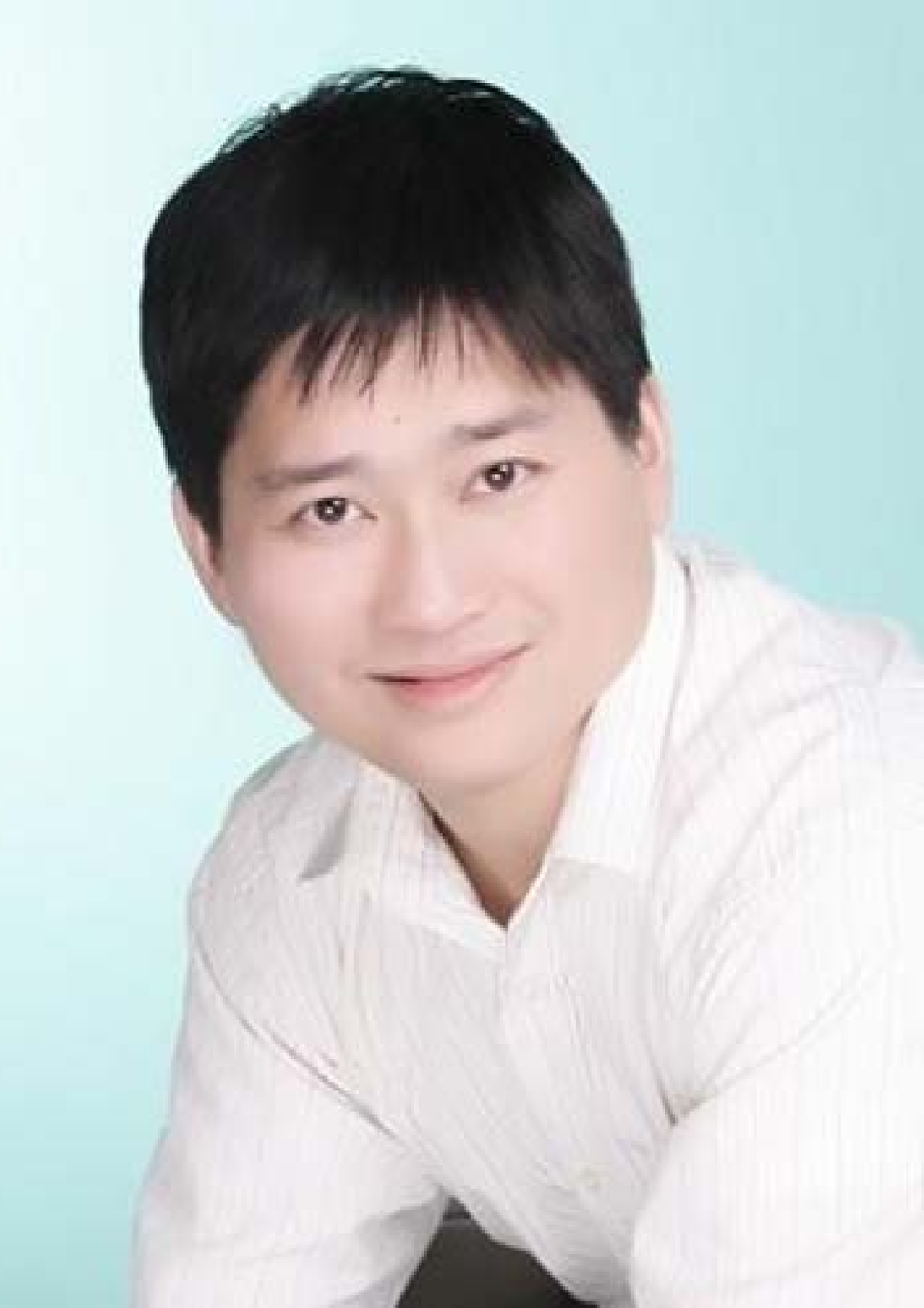}}]{Kaiqi Huang} received his BSc and MSc from Nanjing University of Science Technology, China and obtained his PhD degree from Southeast University. He has worked in National Lab. of Pattern Recognition (NLPR), Institute of Automation, Chinese Academy of Science, China and now he is a professor in NLPR. His current research interests include visual surveillance, digital image processing, pattern recognition and biological based vision and so on. He is a senior member of the IEEE.
\end{IEEEbiography}

\begin{IEEEbiography}[{\includegraphics[width=1.25in,height=1.25in,clip,keepaspectratio]{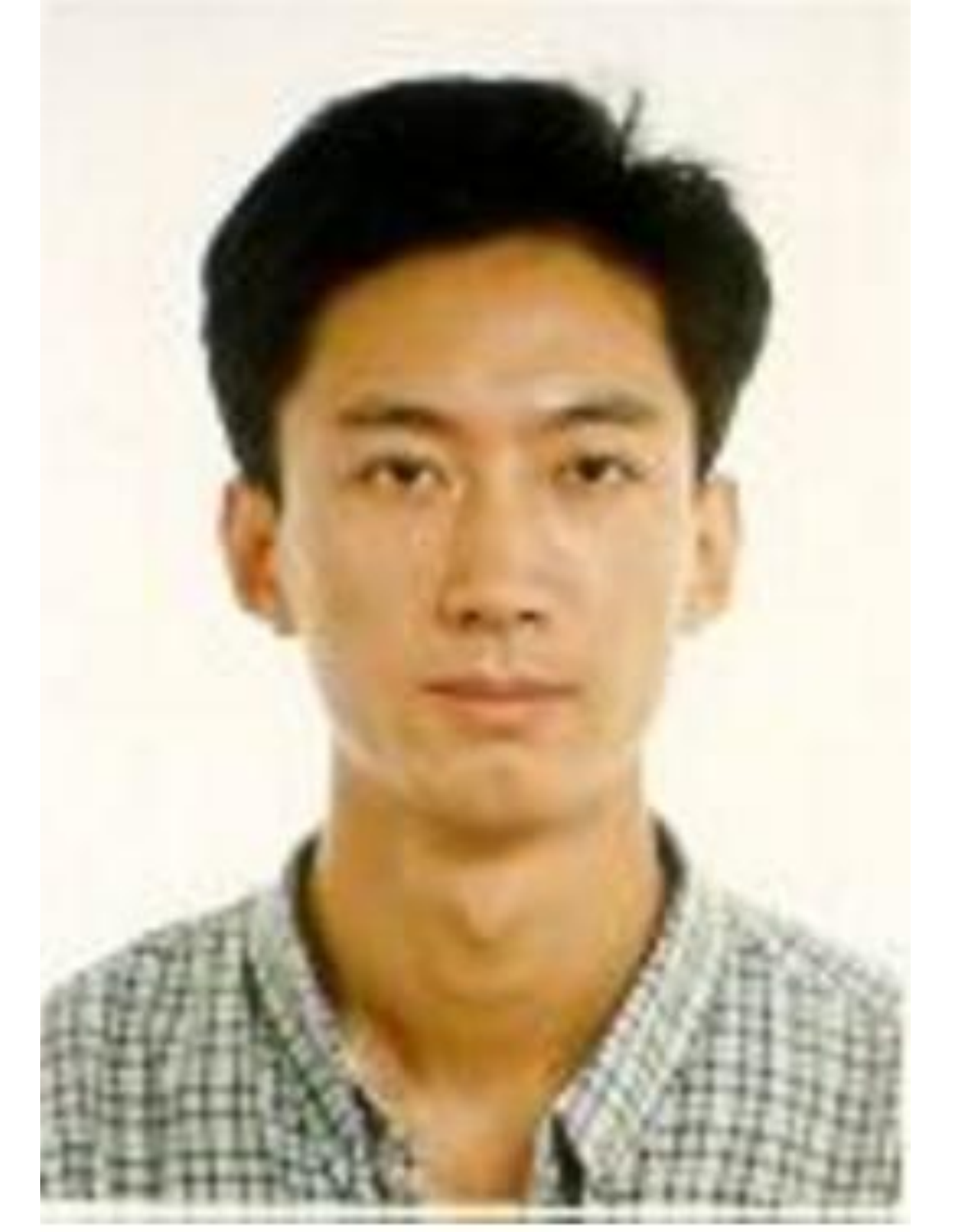}}]{Jianguo Zhang} is currently a Reader of Visual Computation at University of Dundee, UK. He received a PhD in National Lab of Pattern Recognition, Institute of Automation, Chinese Academy of Sciences, Beijing, China, 2002. His research interests include visual surveillance, object recognition, image processing, medical image analysis and machine learning. He is a senior member of the IEEE.
\end{IEEEbiography}

\clearpage
\appendices
\section{Experimental setup in intra-dataset detection task on ILSVRC2013}
\label{sec:appeimagenet}
In the intra-dataset detection task on ILSVRC2013, as we described in Section 5.2, the detection methods (LSSOD \cite{tang2016large} \& our method) are trained on \textit{train} and \textit{val1} sets and are evaluated on \textit{val2} set. However, the training images actually used in our method are slightly different from the ones used in LSSOD. We will clarify the details in this section.

The training process of the intra-dataset detection contains two individual parts, 1) fully supervised training and 2) weakly supervised training. In LSSOD, the fully supervised training corresponds to \textbf{the training of detectors on the first 100 categories} and the weakly supervised training corresponds to \textbf{the training of classifiers on all the 200 categories}. In our method, the fully supervised training and weakly supervised training correspond to \textbf{the training of objectness models} (Section 4.1) and \textbf{the training of objectness-aware models} (Section 4.2) respectively. In both training processes, our experimental setup differs from LSSOD.

\textbf{For the fully supervised training}, LSSOD and our method both utilize the images with bounding box annotations from the first 100 categories in \textit{train} and \textit{val1}. For fully labeled images in \textit{val1}, both the two methods use selective search windows \cite{uijlings2013selective} and ground truth boxes in these images. However, for fully labeled images in \textit{train}, the images are partly labeled where most but not all instances are annotated. Thus, LSSOD only selects 1,000 ground truth boxes per category from \textit{train} and does \textbf{not} utilize the selective search windows in these images (following the same protocol of R-CNN \cite{girshick2014rich}). This manner is suitable for R-CNN, but not for Fast R-CNN \cite{girshick2015fast} framework used in our method. The Fast R-CNN uses the \textit{hierarchical sampling} strategy by first sampling $N$ images and then sampling $R/N$ proposals from each image. If we use the same setting as LSSOD that only utilizes the ground truth boxes from images in \textit{train}, the mini-batch may contain very few proposals (\eg only two proposals when $N$=2 and each image contains only one ground truth box.). To avoid this, for the partly labeled images in \textit{train}, we also utilize both selective search windows and ground truth boxes in these images just as the utilization of \textit{val1} images. That is, for images of the first 100 categories in our \textit{trainval1} split, we utilize both selective search windows and ground truth boxes for fully supervised training. This manner may lead to incorrect annotations but will not harm the final detection performance according to our experimental verification. We train Fast R-CNN object detectors with training images in our \textit{trainval1} split and finally obtain 30.12\% on the first 100 categories. This result is similar to the one (29.72\%) reported by LSSOD when LSSOD trains R-CNN detectors with its settings. The summary of the training data used in fully supervised training are shown in Table \ref{tab:fully}.


\textbf{For the weakly supervised training}, to balance the categories, LSSOD manually selects 1,000 weakly labeled images per category from the \textit{train} set. In contrast, we abandon this manually selection process and directly use the images of the last 100 categories in our \textit{trainval1} set (57,584 images, the last 100 categories) to train objectness-aware detection models. Notice that a portion of \textit{train} images are annotated with image category labels only. In LSSOD, these images are also used for weakly supervised training, but in our method, we do not include these images in our \textit{trainval1} set. Thus the weakly labeled images used in our method are nearly half as much as LSSOD ($\sim$575 images per category \vs 1,000 images per category). The summary of the training data used in fully supervised training are shown in Table \ref{tab:weakly}.

\begin{table*}[]
\renewcommand{\arraystretch}{1.5}
\caption{\label{tab:fully}The training data used in \textbf{fully supervised training} for LSSOD and our robust objectness transfer approach for MSD. For both methods, we use the images belonging to \textbf{the first 100 categories} for training. \textbf{Positive images \& Negative images}: the \textit{train} set contains \textit{negative} images that do not belong to any category of ILSVRC2013 detection dataset. \textbf{Bounding box anno. \& Image cls label}: a portion of positive \textit{train} images are annotated with image category labels only; the other images are annotated with bounding box annotations. \textbf{GT boxes \& SS windows}: ground truth boxes \& selective search windows.}
\small
\begin{center}
\begin{tabular}{|c|c|c|c|l|c|l|l|l|c|l|l|l|c|l|l|l|} \hline
\multirow{4}{*}{Data splits in ILSVRC2013} & \multicolumn{8}{c|}{\begin{tabular}[c]{@{}c@{}}The \textit{train} set \\ (395,909 images in total)\end{tabular}}
                                          & \multicolumn{8}{c|}{\begin{tabular}[c]{@{}c@{}}The \textit{val1} set \\ (10,204 images in total)\end{tabular}}   \\ \cline{2-17}
& \multicolumn{4}{c|}{Positive images}    & \multicolumn{4}{c|}{\multirow{3}{*}{Negative images}} & \multicolumn{8}{c|}{Positive images}
                                          \\ \cline{2-5} \cline{10-17}
& \multicolumn{2}{c|}{Bounding box anno.} & \multicolumn{2}{c|}{\multirow{2}{*}{Image cls label}} & \multicolumn{4}{c|}{}
& \multicolumn{8}{c|}{Bounding box anno.}              \\ \cline{2-3} \cline{10-17}
& GT boxes    & SS windows   & \multicolumn{2}{c|}{}   & \multicolumn{4}{c|}{}& \multicolumn{4}{c|}{GT boxes} & \multicolumn{4}{c|}{SS windows} \\ \hline
\begin{tabular}[c]{@{}c@{}}LSSOD \\ (The training of detectors)\end{tabular}                    & \Checkmark      & \XSolidBrush
& \multicolumn{2}{c|}{\XSolidBrush}   & \multicolumn{4}{c|}{\XSolidBrush}      & \multicolumn{4}{c|}{\Checkmark}  & \multicolumn{4}{c|}{\Checkmark}  \\ \hline
\begin{tabular}[c]{@{}c@{}}Ours MSD \\ (The training of objectness models)\end{tabular}      & \Checkmark      & \Checkmark
& \multicolumn{2}{c|}{\XSolidBrush}   & \multicolumn{4}{c|}{\XSolidBrush}      & \multicolumn{4}{c|}{\Checkmark}  & \multicolumn{4}{c|}{\Checkmark}  \\ \hline
\end{tabular}
\end{center}
\end{table*}

\begin{table*}[]
\renewcommand{\arraystretch}{1.5}
\caption{\label{tab:weakly}The training data used in \textbf{weakly supervised training} for LSSOD and our robust objectness transfer approach for MSD. For LSSOD, we use the images belonging to \textbf{all the 200 categories} for training; for our method, we only use the images of \textbf{the last 100 categories}. \textbf{Bounding box anno.}: the images with bounding box annotations are used for training. It should be noticed that, for these images, only image category labels are utilized and we do not have access to their exact bounding box annotations.}
\small
\begin{center}
\begin{tabular}{|c|c|c|c|l|c|l|l|l|} \hline
\multirow{3}{*}{Data splits in ILSVRC2013} & \multicolumn{4}{c|}{\begin{tabular}[c]{@{}c@{}}The \textit{train} set\\ (395,909 images in total)\end{tabular}}
                                          & \multicolumn{4}{c|}{\begin{tabular}[c]{@{}c@{}}The \textit{val1} set\\ (10,204 images in total)\end{tabular}}
                                          \\ \cline{2-9}
& \multicolumn{2}{c|}{Positive images}    & \multicolumn{2}{c|}{\multirow{2}{*}{Negative images}} & \multicolumn{4}{c|}{Positive images}
                                          \\ \cline{2-3} \cline{6-9}
& Bounding box anno.  & Image cls label   & \multicolumn{2}{c|}{}                                 & \multicolumn{4}{c|}{Bounding box anno.}    \\ \hline
\begin{tabular}[c]{@{}c@{}}LSSOD\\ (The training of classifiers)\end{tabular}
& \Checkmark          & \Checkmark        & \multicolumn{2}{c|}{\XSolidBrush}                     & \multicolumn{4}{c|}{\XSolidBrush}          \\ \hline
\begin{tabular}[c]{@{}c@{}}Ours MSD\\ (The training of objectness-aware models)\end{tabular}
& \Checkmark          & \XSolidBrush      & \multicolumn{2}{c|}{\XSolidBrush}                     & \multicolumn{4}{c|}{\Checkmark}            \\ \hline
\end{tabular}
\end{center}
\end{table*}

In a word, our experimental settings are more suitable for Fast R-CNN framework and avoid the manually selection process used in LSSOD. Our \textit{trainval1} split will be publicly available soon. Meanwhile, we believe that LSSOD will not obtain higher results under our settings, since the amount of weakly labeled images to train classifiers are smaller than the ones in LSSOD and the categories are not balanced in our settings. Thus, we still compare the original LSSOD results in our paper.

\section{Fast R-CNN \vs R-CNN (+LSSOD)}
Notice that LSSOD \cite{tang2016large} adopts the R-CNN framework, which is different from our proposed method that applies the Fast R-CNN framework. Thus to ensure the fairness, we aim to reproduce LSSOD based on the Fast R-CNN framework to perform the comparison. The training process of LSSOD contains two steps, 1) the training of classifiers and 2) the training of object detectors. To reproduce LSSOD with Fast R-CNN, we aim to utilize Fast R-CNN models in both steps.

In the first step, to train classifiers with Fast R-CNN models, we use the ROI-pooling layer for the entire image, that is, to feed the ROI-pooling layer with the \textit{original image size}, rather than the \textit{proposal size}, to compute the features for the whole image. An additional detail to mention here is that we randomly select a smaller size from the original image while using the ROI-pooling layer to serve as the widely used ``random cropping" for data augmentation. In this step, our Fast R-CNN classifiers are fine-tuned from pre-trained ImageNet AlexNet models \cite{krizhevsky2012imagenet}. Then, in the second step, we continue to train Fast R-CNN object detectors initialized from the Fast R-CNN classifiers obtained in step one. Finally, we calculate the classifier and detector differences and then adapt the differences to the weakly labeled categories following the same setting described in LSSOD.

However, using this strategy, we obtain a surprising result that the performance of LSSOD \textbf{decreased} dramatically (\textbf{from 19.02\% to 12.49\%}) when applying the Fast R-CNN framework. We have also attempted other strategies to reproduce LSSOD, such as a) training R-CNN classifiers in step one and fine-tuning R-CNN classifiers to Fast R-CNN detectors in step two or b) using an \textit{inverse process} that first training Fast R-CNN detectors and then fine-tuning detectors for classification. According to our experimental verification, all these strategies cannot get better results, and \textbf{12.49\% is already the best result} of LSSOD with Fast R-CNN framework.

We explore the reasons from the training process of LSSOD: in LSSOD, both classifiers and object detectors are trained with 8-layer AlexNet models, and LSSOD computes \textbf{the classifier and detector differences} by directly \textbf{subtracting} the parameters of classifier models from the ones of object detector models. Then LSSOD transfers the ``differences'' from strong categories to weak categories following some pre-designed strategies. Finally, LSSOD adapts the classifier of a weak category to the required detector by directly \textbf{adding} these transferred ``differences'' to the parameters of its classifier.

\begin{table*}[]
\renewcommand{\arraystretch}{1.5}
\caption{\label{tab:compare}Comparisons of LSSOD with the Fast R-CNN framework and the R-CNN framework. The classifier and object detectors obtain similar results, but the transfer models trained with Fast R-CNN obtain performance relatively lower than the original ones trained with R-CNN. }
\begin{center}
\begin{tabular}{|l|l|l|}
\hline   Method   & \begin{tabular}[c]{@{}c@{}}LSSOD\\(Fast R-CNN)\end{tabular}  & \begin{tabular}[c]{@{}c@{}}LSSOD\\(R-CNN)\end{tabular}\\ \hline
Classifiers (evaluated on weak categories)                      & \multicolumn{1}{c|}{9.72}    & \multicolumn{1}{c|}{10.31 \cite{tang2016large}} \\
Detectors (evaluated on strong categories)                      & \multicolumn{1}{c|}{30.12}    & \multicolumn{1}{c|}{29.72 \cite{tang2016large}} \\ \hline
                                                                                                                                       \hline
Transfer models (evaluated on weak categories)                  & \multicolumn{1}{c|}{12.49 $\downarrow$}    & \multicolumn{1}{c|}{19.02 \cite{tang2016large}} \\ \hline

\end{tabular}
\end{center}
\end{table*}

\begin{table*}[]
\renewcommand{\arraystretch}{1.5}
\caption{\label{tab:voc-hard}Object detection performance (mAP) on PASCAL VOC 2007 \textit{test} for different training strategies. ``XX-hard'' indicates the method that applies hard negative mining strategy to XX. All the four methods are trained with AlexNet. }
\begin{center}
\resizebox{\textwidth}{!} {
\begin{tabular}{|l|llllllllllllllllllll|l|}
\hline
Method & \multicolumn{1}{c}{aero} & \multicolumn{1}{c}{bike} & \multicolumn{1}{c}{bird} & \multicolumn{1}{c}{boat} & \multicolumn{1}{c}{bottle} & \multicolumn{1}{c}{bus} &
\multicolumn{1}{c}{car} & \multicolumn{1}{c}{cat} & \multicolumn{1}{c}{chair} & \multicolumn{1}{c}{cow} & \multicolumn{1}{c}{table} & \multicolumn{1}{c}{dog} & \multicolumn{1}{c}{horse} & \multicolumn{1}{c}{mbike} & \multicolumn{1}{c}{person} & \multicolumn{1}{c}{plant} & \multicolumn{1}{c}{sheep} & \multicolumn{1}{c}{sofa} & \multicolumn{1}{c}{train} & \multicolumn{1}{c|}{tv} & \multicolumn{1}{c|}{mAP} \\ \hline
                                                                                  \hline
B-WSD             & \multicolumn{1}{c}{40.5} & \multicolumn{1}{c}{35.3} & \multicolumn{1}{c}{19.5} & \multicolumn{1}{c}{5.8}  & \multicolumn{1}{c}{7.7}
                  & \multicolumn{1}{c}{38.9} & \multicolumn{1}{c}{39.9} & \multicolumn{1}{c}{23.3} & \multicolumn{1}{c}{1.6}  & \multicolumn{1}{c}{25.0}
                  & \multicolumn{1}{c}{11.1} & \multicolumn{1}{c}{25.2} & \multicolumn{1}{c}{29.9} & \multicolumn{1}{c}{49.5} & \multicolumn{1}{c}{21.3}
                  & \multicolumn{1}{c}{16.4} & \multicolumn{1}{c}{24.4} & \multicolumn{1}{c}{16.8} & \multicolumn{1}{c}{35.1} & \multicolumn{1}{c|}{10.5}
                  & \multicolumn{1}{c|}{23.87} \\
B-WSD-hard             & \multicolumn{1}{c}{39.1} & \multicolumn{1}{c}{36.8} & \multicolumn{1}{c}{24.3} & \multicolumn{1}{c}{5.7}  & \multicolumn{1}{c}{11.0}
                  & \multicolumn{1}{c}{41.7} & \multicolumn{1}{c}{39.5} & \multicolumn{1}{c}{21.0} & \multicolumn{1}{c}{1.9}  & \multicolumn{1}{c}{30.4}
                  & \multicolumn{1}{c}{11.8} & \multicolumn{1}{c}{22.4} & \multicolumn{1}{c}{33.8} & \multicolumn{1}{c}{48.9} & \multicolumn{1}{c}{21.5}
                  & \multicolumn{1}{c}{17.2} & \multicolumn{1}{c}{28.8} & \multicolumn{1}{c}{16.2} & \multicolumn{1}{c}{36.8} & \multicolumn{1}{c|}{10.9}
                  & \multicolumn{1}{c|}{24.98} \\ \hline \hline
Ours-MSD  & \multicolumn{1}{c}{55.8} & \multicolumn{1}{c}{56.6} & \multicolumn{1}{c}{41.1} & \multicolumn{1}{c}{35.1} & \multicolumn{1}{c}{22.8}
                  & \multicolumn{1}{c}{60.1} & \multicolumn{1}{c}{58.5} & \multicolumn{1}{c}{55.0} & \multicolumn{1}{c}{10.3} & \multicolumn{1}{c}{48.5}
                  & \multicolumn{1}{c}{22.2} & \multicolumn{1}{c}{50.5} & \multicolumn{1}{c}{55.8} & \multicolumn{1}{c}{61.6} & \multicolumn{1}{c}{12.8}
                  & \multicolumn{1}{c}{21.7} & \multicolumn{1}{c}{44.4} & \multicolumn{1}{c}{26.1} & \multicolumn{1}{c}{46.8} & \multicolumn{1}{c|}{49.4}
                  & \multicolumn{1}{c|}{41.77}  \\
Ours-MSD-hard  & \multicolumn{1}{c}{54.9} & \multicolumn{1}{c}{54.7} & \multicolumn{1}{c}{40.5} & \multicolumn{1}{c}{38.6} & \multicolumn{1}{c}{22.3}
                  & \multicolumn{1}{c}{60.9} & \multicolumn{1}{c}{57.9} & \multicolumn{1}{c}{57.2} & \multicolumn{1}{c}{9.7} & \multicolumn{1}{c}{51.4}
                  & \multicolumn{1}{c}{22.0} & \multicolumn{1}{c}{52.2} & \multicolumn{1}{c}{61.6} & \multicolumn{1}{c}{63.2} & \multicolumn{1}{c}{10.1}
                  & \multicolumn{1}{c}{17.3} & \multicolumn{1}{c}{47.8} & \multicolumn{1}{c}{25.2} & \multicolumn{1}{c}{51.9} & \multicolumn{1}{c|}{51.2}
                  & \multicolumn{1}{c|}{42.54}  \\ \hline

\end{tabular}
}
\end{center}
\end{table*}

In R-CNN framework, training classifiers can be viewed as a special case of training object detectors. Both images (used for training classifiers) and region proposals (used for training detectors) are resized to $224\times224$ and R-CNN performs a complete ConvNet forward pass (from conv1 to fc8) for them. Thus, in R-CNN, classifier models and object detector models are \textbf{compatible} and such \textit{subtraction} operator is appropriate to get the ``differences''. However, in Fast R-CNN framework, the classifiers and object detectors are \textbf{incompatible}. When training classifiers, Fast R-CNN still performs a complete ConvNet forward pass for images, that is, Fast R-CNN classifiers compute image-level features from original images (data layer); by contrast, when training object detectors, Fast R-CNN crops region-level features from internal convolutional layers (conv5 layer). \textbf{In Fast R-CNN, due to such ``mismatch'' between classifiers and object detectors, it is very difficult to compute the reliable ``differences'' by simply subtracting the parameters between the two CNN models, and then the good-performance transfer models (object detectors on weak categories) cannot be obtained by directly adding such unreliable ``differences''}. To support our assumption, we respectively use Fast R-CNN classifiers, Fast R-CNN detectors and Fast R-CNN transfer models obtained in the training of LSSOD for detection. The evaluation results are shown in Table \ref{tab:compare} (mAP), and we can see that:

\begin{itemize}
\item{When Fast R-CNN classifier is directly used for detection, it obtains 9.72\% on weak categories (the last 100 categories), which is similar to LSSOD (R-CNN) (10.31\%);}
\item{The Fast R-CNN detector reaches 30.12\% on strong categories (the first 100 categories), which is also similar to LSSOD (R-CNN) (29.72\%);}
\item{The final transfer model obtained by adapting the classifier and object detector differences cannot obtain comparable results with LSSOD (R-CNN) (12.49\% \vs 19.02\%).}
\end{itemize}
These results verify our assumption, and based on these considerations, we compare our method with the original LSSOD (R-CNN) in our paper.

\section{Hard negative mining in weakly/mixed supervised detection}
In early fully supervised detection works that are based on standard multiple instance learning (MIL), \eg, DPM \cite{felzenszwalb2010object}, hard negative mining is a standard strategy to address the issues caused by imbalanced categories. In standard MIL setting, \textit{all} regions in negative images are considered as negative samples and it will lead to extremely imbalanced positive to negative ratio. Thus, the hard negative mining is necessary in these methods.

Some of the weakly supervised detection (WSD) methods, \eg, Multi-fold \cite{cinbis2017weakly}, are also based on MIL framework. In Multi-fold, the MIL setting has been modified to only consider \textit{top-scored} regions in negative images as negative samples. In this case, the ratio of positives to negatives could be largely reduced. However, due to the class-imbalance problem existing in PASCAL VOC dataset, the hard negative mining is still utilized to improve the performance in Multi-fold.

Recent WSD methods, \eg, WSDDN \cite{bilen2016weakly}, do not apply the hard negative mining strategy in their experiments. Actually, in WSDDN, the second branch tends to capture only one simplest negative sample with \textit{softmax} operation. Thus, for a fair comparison with these WSD methods, we also do not utilize hard negative mining for the proposed MSD method (Ours-MSD) and other baseline methods (\eg, B-WSD) in our settings.

Recently, OHEM \cite{shrivastava2016training} has verified the effectiveness of the hard negative mining in fully supervised detection. Thus, we believe that it is a worthy experiment to test the effect of the hard negative mining in weakly/mixed supervised detection tasks. We respectively conduct experiments on B-WSD and Ours-MSD. The detection models are trained with AlexNet.

Concretely, we first train the weakly/mixed supervised detector for B-WSD/Ours-MSD following the same settings in Section 5.3.2. Then we apply the hard negative mining strategy to the obtained detector as follows:

1) for a category, \eg, \textit{cat}, the \textit{cat} scores of all regions in all training images can be obtained using the learned detector.

2) for each training image, the region with highest \textit{cat} score is selected. The selected regions belonging to \textit{cat} images are positive \textit{cat} regions; the selected regions of other images (\ie, images that do not contain \textit{cat}
) are negative regions.

3) the negative \textit{cat} regions are sorted according to their \textit{cat} scores. Then the negative regions with higher \textit{cat} scores are selected as hard negatives. The ratio of positive regions to hard negative regions is controlled to 1:3.

4) Repeat 1) $\sim$ 3) to select hard negative regions for all categories. Using the selected regions, the obtained detector is fine tuned for 10 epochs with fixed learning rate (\ie, $5\times{10}^{-6}$).

The results are shown in Table \ref{tab:voc-hard}. It can be seen that for both low-performance weakly supervised detector (B-WSD) and high-performance mixed supervised detector (Ours-MSD), applying hard negative mining improves the detection performance (24.98\% \vs 23.87\%, 42.54\% \vs. 41.77\%), which confirms the effectiveness of the hard negative mining in detection approach.


\end{document}